\documentclass[10pt,journal,compsoc]{IEEEtran}
\ifCLASSOPTIONcompsoc
\usepackage[nocompress]{cite}
\else
\usepackage{cite}
\fi
\ifCLASSINFOpdf
\else
\fi
\usepackage{url}

\usepackage{amsmath}
\usepackage{amsfonts}
\usepackage{graphicx}
\usepackage{subfigure}
\usepackage{booktabs}
\usepackage{multirow}
\usepackage{amsmath}
\usepackage{dsfont}
\usepackage{xcolor}

\begin{document}
	%
	\title{A Clustering Framework for Unsupervised and Semi-supervised New Intent Discovery}
	%
	%
	%
	%

	\author{Hanlei Zhang, Hua Xu, \textit{Member, IEEE,} Xin Wang, Fei Long, Kai Gao
		\IEEEcompsocitemizethanks{
			\IEEEcompsocthanksitem 
   H. Zhang, H. Xu, X. Wang, and F. Long are with State Key Laboratory of Intelligent Technology and Systems, Department of Computer Science and Technology, 
			Tsinghua University, Beijing 100084, China.\protect\\	
			E-mail: zhang-hl20@mails.tsinghua.edu.cn; xuhua@tsinghua.edu.cn;\protect\\wx\_hebust@163.com; long-f20@mails.tsinghua.edu.cn
			\IEEEcompsocthanksitem X. Wang and K. Gao are with  
			School of Information Science and Engineering, Hebei University of Science and Technology, Shijiazhuang 050018, China. 
			Email: wx\_hebust@163.com; gaokai@hebust.edu.cn 
			\IEEEcompsocthanksitem Hua Xu is the corresponding author. Part of the research was completed in cooperation with Samton (Jiangxi) Technology Development Co., Ltd. 
   \IEEEcompsocthanksitem This work was supported by the National Natural Science Foundation of China (Grant No. 62173195), the National Science and Technology Major Project towards the new generation of broadband wireless mobile communication networks of Jiangxi Province (03 and 5G Major Project of Jiangxi Province) (Grant No. 20232ABC03402), High-level Scientific and Technological Innovation Talents "Double Hundred Plan" of Nanchang City in 2022 (Grant No. Hongke Zi (2022) 321-16), and Natural Science Foundation of Hebei Province, China
   (Grant No. F2022208006). 
   \IEEEcompsocthanksitem Data and codes are available at~\protect\url{https://github.com/thuiar/TEXTOIR}.
	}}

\IEEEtitleabstractindextext{%
    \begin{abstract}
     New intent discovery is of great value to natural language processing, allowing for a better understanding of user needs and providing friendly services. However, most existing methods struggle to capture the complicated semantics of discrete text representations when limited or no prior knowledge of labeled data is available. To tackle this problem, we propose a novel clustering framework, USNID, for \textbf{u}nsupervised and \textbf{s}emi-supervised \textbf{n}ew \textbf{i}ntent \textbf{d}iscovery, which has three key technologies. First, it fully utilizes unsupervised or semi-supervised data to mine shallow semantic similarity relations and provide well-initialized representations for clustering. Second, it designs a centroid-guided clustering mechanism to address the issue of cluster allocation inconsistency  and provide high-quality self-supervised targets for representation learning. Third, it captures high-level semantics in unsupervised or semi-supervised data to discover fine-grained intent-wise clusters by optimizing both cluster-level and instance-level objectives. We also propose an effective method for estimating the cluster number in open-world scenarios without knowing the number of new intents beforehand. USNID performs exceptionally well on several benchmark intent datasets, achieving new state-of-the-art results in unsupervised and semi-supervised new intent discovery and demonstrating robust performance with different cluster numbers.	
\end{abstract}
		
		}

	\maketitle

	\IEEEdisplaynontitleabstractindextext

	%
	\IEEEpeerreviewmaketitle

	\IEEEraisesectionheading{\section{Introduction}\label{sec:introduction}}

	%
	%
	%
	%
	\IEEEPARstart{D}{iscovering} new intents is an important aspect of natural language processing, as it has numerous applications in dialogue and user-modeling systems~\cite{ijcai2021p622,li2022automatically}. These newly discovered intents can help to enrich the intent taxonomy and improve the natural language understanding capabilities of dialogue systems in interacting with users~\cite{lin2020discovering}. In addition, they can be used to improve user profiles and analyze user interests and preferences, leading to more personalized services~\cite{9319534}. 
\begin{figure*}
    \centering
    \includegraphics[scale=.53]{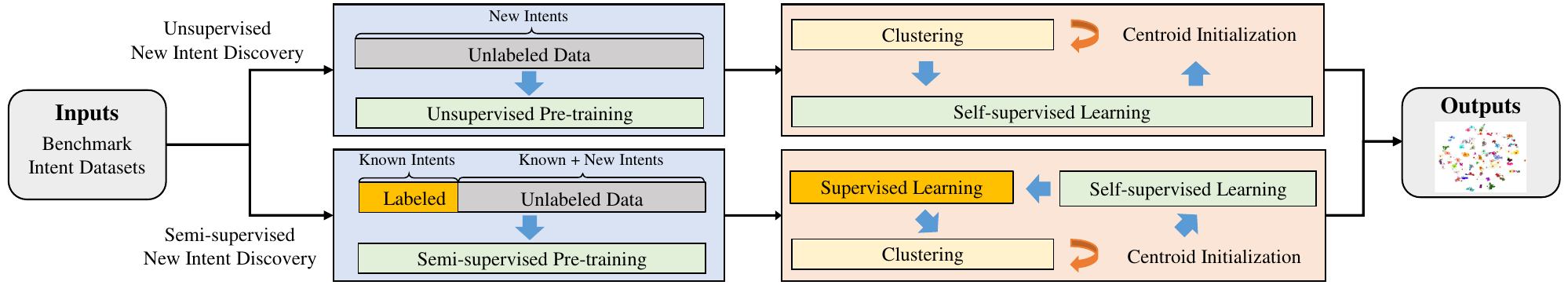}
    \caption{Overview of the USNID framework for new intent discovery. The unsupervised pipeline first captures primary semantic features from unlabeled data through a pre-training phase and then learns high-level intent representations with two iterative steps. One uses cluster centroids as guidance to obtain consistent targets aligned with the last clustering. The other uses those targets to learn friendly representations for the next clustering. The semi-supervised pipeline further leverages the labeled data as prior knowledge  to improve clustering and representation learning.}
    \label{overall_framework}
\end{figure*}

 Typical intent understanding tasks use annotated intent corpora to train a supervised classification model~\cite{schuurmans2019intent}, with the goal of accurately predicting the corresponding intent category for each text utterance. However, in widespread real-world applications, there are two main difficulties. First, pre-defined intent categories may not be sufficient to capture the complexity and diversity of user needs, requiring the effective mining of potential clusters of user demands and the formation of new intents. Second, in practice, there is often a large amount of unlabeled data, making it labor-intensive and time-consuming to annotate a sufficient quantity of high-quality intent data. Therefore, it is of great significance to find ways to make full use of unlabeled data or semi-supervised data with a limited amount of labels.
 
 To address these issues, we consider the new intent discovery task,  which is  a clustering problem. For semi-supervised new intent discovery, we randomly select a portion of intent classes as known and the rest as new intents. Considering the scarcity of labeled data in real applications, we mask most labels with known intents (i.e., 90\%). The masked known-intent samples and new-intent samples constitute the unlabeled data. The goal is to use limited labeled and a large amount of unlabeled data to find known and discover new intent groups. For unsupervised new intent discovery, it aims to discover new intent groups without any prior knowledge of labeled data.

Novel category discovery (NCD)~\cite{han2021autonovel,fini2021unified} is similar to our task in computer vision (CV). The main difference is that it assumes unlabeled data only come from novel classes, which is inapplicable in real-world scenarios as unlabeled data usually contain a mix of known and novel categories. While GCD~\cite{vaze2022generalized} proposed a generalized setting to address this issue, it still requires a larger proportion of labeled data (e.g., 50\% v.s. 10\%) and does not provide a solution for the unsupervised setting. Additionally, experiments show that the methods used in NCD~\cite{Han2019learning} and GCD~\cite{vaze2022generalized}  have limitations when applied to our task due to their difficulty in learning the complex semantics of discrete text representations. 

The study of new intent discovery has gained attention in recent years. We made the first trials on this task~\cite{lin2020discovering, Zhang_Xu_Lin_Lyu_2021}, and subsequent works have made further progress in improving performance~\cite{kumar-etal-2022-intent,wei2022semi,zhang2022new}. It has also been successfully applied in real applications to discover user consumption intents~\cite{li2022automatically}. 
There are three main challenges in this task. Firstly, current methods still heavily rely on labeled data, and their performance suffers significantly in a completely unsupervised setting without any extra knowledge~\cite{Zhang_Xu_Lin_Lyu_2021,zhang2022new}. Secondly, in a semi-supervised scenario, we need to take full advantage of limited labeled data and transfer its knowledge to guide unlabeled data to learn intent representations conducive to clustering. Thirdly, the number of new intents may not be known in advance. In this case, effectively estimating the cluster number is also a crucial factor in determining the final performance.

To tackle these problems, we propose a novel clustering framework called USNID for unsupervised and semi-supervised new intent discovery, as shown in Figure~\ref{overall_framework}. The unsupervised new intent discovery consists of two key steps. The first step is to pre-train the model by applying unsupervised contrastive learning on unlabeled data. We construct positive pairs with each sample and its corresponding strong data augmentation. The second step is to learn high-level intent-wise characteristics through an iterative process of clustering and self-supervised learning.

However, the cluster assignments from the partition-based method (e.g., K-Means~\cite{macqueen1967some}) may not be consistent for the same sample across different clustering, making it difficult to use them as pseudo-labels to train a stable classifier for discriminating new intent classes. To overcome this issue, we introduce a centroid-guided clustering mechanism that leverages cluster centroids from adjacent clustering as guidance to obtain aligned targets. One way to achieve this is by minimizing Euclidean distances between the two cluster centroid matrices globally to obtain an alignment projection. Still, each clustering process can still be inefficient and prone to falling into local optima due to the randomness of initial cluster centroid selection. To mitigate this, we propose a centroid initialization strategy that leverages the cluster centroids from the previous iteration's clustering to initialize the current iteration's clustering. This strategy can improve convergence with the prior knowledge of historical clustering information. Moreover, the produced cluster assignments are usually consistent with the results of centroid alignment, which can be directly used as self-supervised signals for representation learning. It is also important to select a suitable self-supervised learning objective to provide friendly representations for the next clustering. The designed objective captures both cluster-level and instance-level information using aligned targets. The former learns a discriminator to distinguish different fine-grained intent classes, while the latter aims to enhance the semantic similarity relationships between instances with intra-class compactness and inter-class separation properties. 

The semi-supervised new intent discovery process uses labeled data in two ways to improve performance. First, we optimize the pre-training phase using a combination of semi-supervised contrastive learning and known-intent classification objectives, which utilize limited labeled data to guide the learning of primary semantic features in a large amount of unlabeled data. Second, we incorporate a supervised contrastive learning objective to enhance the memory of the limited labeled data 
and improve the ability to cluster and learn representations. Nevertheless, the approach still requires the cluster number to be specified in advance, which is not practical in real-world situations. Thus, we propose a simple and effective method for estimating the number of new-intent classes. Our method only requires one clustering operation using a large, pre-defined number of clusters. The main idea is to use the knowledge acquired during the pre-training phase to find high-quality clusters that are denser than a certain threshold. In semi-supervised scenarios, limited labeled data can be used to induce clusters corresponding to known intents and avoid interference with the estimation of the number of new intents.

Our USNID framework is evaluated on several benchmark intent datasets and compared with 15 algorithms that can be used in unsupervised and semi-supervised new intent discovery. 
It is the first successful attempt at unsupervised new intent discovery, resulting in an absolute increase of 20-30$\%$ adjusted rand index (ARI) over the state-of-the-art (SOTA) unsupervised clustering method. In addition, it achieves new SOTA performance in semi-supervised new intent discovery, showing substantial improvements over previous best-performing methods under different known class ratios. The method of estimating the number of intent classes is also evaluated and found to accurately predict the actual number with the lowest errors compared with other estimation methods. Even when the cluster number is varied in a wide range, our approach still achieves robust and the best performance among all methods. 

\section{Related Works}
In this section, we briefly review the most relevant research in areas of unsupervised and constrained clustering, novel class discovery, and new intent discovery.

There are numerous classic unsupervised clustering technologies in the literature~\cite{macqueen1967some,gowda1978agglomerative}. K-Means is a particularly attractive partitioning method among them due to its simplicity and relatively low time complexity~\cite{jain1999data}. However, it can suffer from poor performance due to the arbitrary sampling of centroids.  Therefore, several variants of K-Means have been proposed to address this issue~\cite{ruspini1969new,arthur2007k}.  K-Means++~\cite{arthur2007k} is selected in this work due to its superior convergence and speed. Deep neural networks (DNNs) have gained popularity in recent years due to their proficiency in handling high-dimensional data and capture complex underlying semantics~\cite{lecun2015deep}. As a result, deep clustering methods have been widely studied~\cite{ren2022deep}. For example, Deep embedded clustering (DEC)~\cite{xie2016unsupervised} utilizes a stacked autoencoder (SAE)~\cite{vincent2010stacked} for low-dimensional feature learning and cluster assignment optimization.  Deep clustering network (DCN)~\cite{yang2017towards} also uses an SAE, optimizing both reconstruction loss and K-Means-like regularization. Deep adaptive clustering (DAC)~\cite{chang2017deep} learns pairwise similarities from confident samples, and DeepCluster~\cite{caron2018deep} alternates between clustering and feature learning. Unsupervised contrastive learning~\cite{chen2020simple} is a rising approach, as seen in Contrastive clustering (CC) that performs instance and cluster-level contrastive learning, and SCCL~\cite{chen2020simple}, which optimizes instance-level contrastive loss and clustering loss for superior text clustering.

Constrained clustering\cite{10.5555/1404506}, introduced to enhance unsupervised clustering through extra supervised signals (e.g., labeled data), includes methods like COP-KMeans~\cite{wagstaff2001constrained} which incorporates hard pairwise constraints. PCK-Means~\cite{basu2004active} handles constraint violations with penalty terms while MPCK-Means~\cite{bilenko2004integrating} adds a distance metric learning objective. DNNs are employed for powerful constrained clustering representations in methods like KCL~\cite{hsu2018learning} and MCL~\cite{hsu2018multiclass}. KCL trains a DNN with pairwise similarity data and generates weak-supervised signals for unlabeled data. MCL uses categorical distribution similarities as weak pairwise constraints. ASFS~\cite{YuSLCHH19} optimizes relationships in diverse data types using unlabeled data and semantic regression, and it uses a graph-based constraint for accurate label prediction. ASTCA~\cite{YuanCLH22} introduces an adaptive model applied successfully in UAV tracking, providing a unique perspective on unsupervised clustering.

Novel class discovery~\cite{han2021autonovel} in computer vision aims to identify new visual classes using labeled data. Approaches include DTC~\cite{Han2019learning}, which pre-trains a model with labeled data and incorporates temporal ensemble predictions into DEC loss, and introduces a method to estimate the number of new classes. RankStats~\cite{han2019automatically} uses unlabeled, self-augmented data in pre-training and calculates pairwise similarities via ranking statistics. UNO~\cite{fini2021unified} optimizes with a unified cross-entropy loss by swapping pseudo-labels of concatenated neural classifier outputs from labeled and unlabeled data. However, these methods assume unlabeled data includes only new classes, which might not be appicable in the real world. GCD~\cite{vaze2022generalized} rectifies this by accommodating both known and new classes in unlabeled data, combining supervised and unsupervised contrastive losses for representation learning and semi-supervised k-means for inference. Despite its superior performance in this task, it struggles with discrete text representations.

The research of new intent discovery is still in its infancy. Prior studies typically focus on known intent classification within closed-world scenarios, utilizing typical intent benchmark datasets~\cite{Casanueva2020,zhang2022mintrec}. More recently, attention has shifted to a related area known as open intent detection~\cite{Zhang_Xu_Lin_2021,10097558}, which seeks to detect the unknown class during testing but lacks the ability to discern fine-grained new classes. We have conducted a pilot study using the CDAC+~\cite{lin2020discovering} algorithm, which first captures pairwise sentence relationships with the guidance of labeled data and then refines cluster assignments with the DEC loss. Another of our works, DeepAligned~\cite{Zhang_Xu_Lin_Lyu_2021}, initializes intent representations under the supervision of labeled data and then iteratively performs clustering and  representation learning, aligning cluster centroids between adjacent iterations to obtain consistent self-supervised signals. DCSC~\cite{wei2022semi} improves the pre-training stage by applying contrastive losses to both labeled and unlabeled data. It mainly uses the SwAV~\cite{caron2020unsupervised} algorithm for unsupervised learning, which requires each sample to predict the swapped view and uses Sinkhorn-Knopp~\cite{cuturi2013sinkhorn} to produce soft cluster assignments. MTP-CLNN~\cite{zhang2022new} is the current SOTA method, which has two key features. First, it enhances representations by incorporating strong prior knowledge from a network pre-trained on external data in the intention domain (i.e., CLINC dataset~\cite{larson-etal-2019-evaluation}) and adding a masked language modeling (MLM) task. Second, it adapts the SCAN algorithm~\cite{van2020scan} to the semi-supervised setting, creating positive pairs with K-nearest neighbors (KNNs) or samples with the same labels for contrastive learning. However, this method relies heavily on the selected external data, and its performance drops dramatically in a purely unsupervised scenario~\cite{zhang2022new}. 

\begin{figure*}[t!]
    \centering  		\includegraphics[width=2.05\columnwidth ]{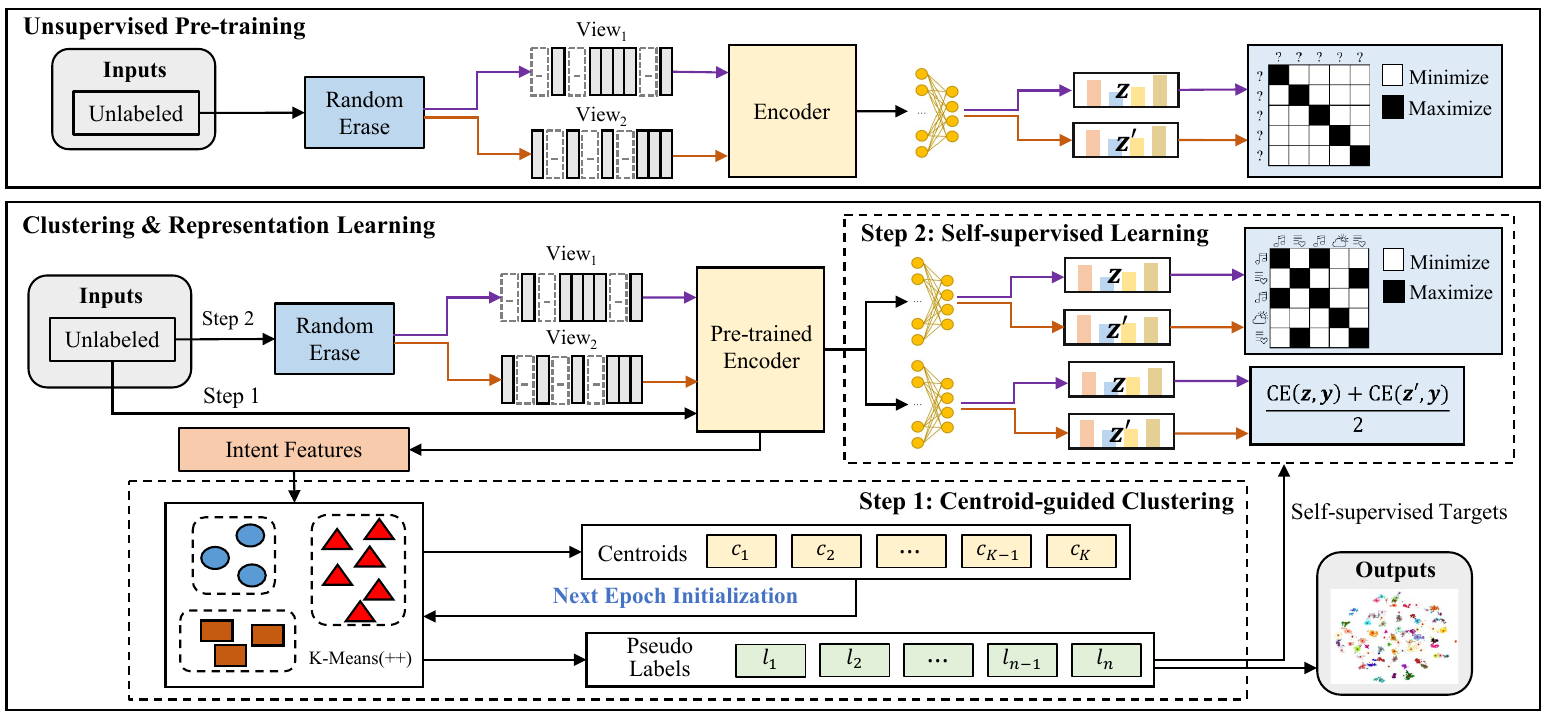}
		\caption{\label{unsup}The pipeline of unsupervised new intent discovery. It first pre-trains the model by applying unsupervised contrastive learning with strong augmented samples. Then, it alternatively performs clustering and representation learning. On the one hand, an efficient centroid-guided clustering algorithm is introduced to produce aligned cluster assignments between adjacent clustering, which can converge well and be used as high-quality self-supervised signals. On the other hand, we learn cluster-level and instance-level information to obtain clustering-friendly intent representations. }
	\end{figure*}  
	\section{Problem Formulation}

\textbf{Unsupervised setting:} We are given an intent dataset $\mathcal{D}_{\textrm{un}}=\{\boldsymbol{x}_{i}|y_{i} \in \mathcal{I}, i=1,..., N\}$, where $\boldsymbol{x}_{i}$ is the $i^{\textrm{th}}$ utterance, $y_{i}$ is the ground-truth label (unseen during training), $N$ is the number of all utterances. $\mathcal{I}=\{\mathcal{I}_i\}^{K}_{i=1}$ is the set of intent labels, where $K$ is the number of intent classes. The goal of unsupervised new intent discovery is to cluster $\{\boldsymbol{x}_{i}\}^{N}_{i=1}$ into $K$ intent groups.

    \textbf{Semi-supervised setting:} We are given an intent dataset $\mathcal{D}_{\textrm{semi}}=\{\mathcal{D}^{l}_{\textrm{semi}},\mathcal{D}^{u}_{\textrm{semi}}\}$, where $\mathcal{D}^{l}_{\textrm{semi}}$ and $\mathcal{D}^{u}_{\textrm{semi}}$  are subsets with limited labeled data (e.g., the labeled ratio $\frac{|\mathcal{D}^{l}_{\textrm{semi}}|}{|\mathcal{D}_{\textrm{semi}}|}<10\%$) and unlabeled data, respectively. 
    
    Specifically, $\mathcal{D}^{l}_{\textrm{semi}}=\{(\boldsymbol{x}_{i}, y_{i})|y_{i}\in \mathcal{I}^{\textrm{known}}, i=1,..., M\}$, where $M$ is the number of labeled utterances, $\mathcal{I}^{\textrm{known}}=\{\mathcal{I}_i\}^{K^{\textrm{known}}}_{i=1}$ is the set of known intent labels. $K^{\textrm{known}}$ is the number of known intent classes which is smaller than $K$ (e.g., the known class ratio $\frac{K^{\textrm{known}}}{K}$ is varied among 25\%, 50\%, and 75\% in this task). 
    
    $\mathcal{D}^{u}_{\textrm{semi}}=\{\boldsymbol{x}_{i}|y_{i}\in \mathcal{I}, i=M+1,..., N\}$, where $\mathcal{I}=\{\mathcal{I}^{\textrm{known}}, \mathcal{I}^{\textrm{new}}\}$, and $\mathcal{I}^{\textrm{new}}=\{\mathcal{I}_i\}^K_{i={K}^{\textrm{known}}+1}$ is the set of new intent labels. Note that $\mathcal{D}^{u}_{\textrm{semi}}$ also contains samples from $\mathcal{I}^{\textrm{known}}$, which is 
   closer to real-world applications with a mixture of both known and new classes for unlabeled data. In comparison,  $\mathcal{D}^{u}_{\textrm{semi}}$ only contains samples from $\mathcal{I}^{\textrm{new}}$ in the similar new class discovery task ~\cite{han2021autonovel}. The goal of semi-supervised new intent discovery is to use $\mathcal{D}^{l}_{\textrm{semi}}$ as prior knowledge to help learn clustering-friendly representations and find known and discover new intent groups. 
   
	\section{Methodologies}
\begin{figure*}[t!]
\centering\includegraphics[width=2.05\columnwidth ]{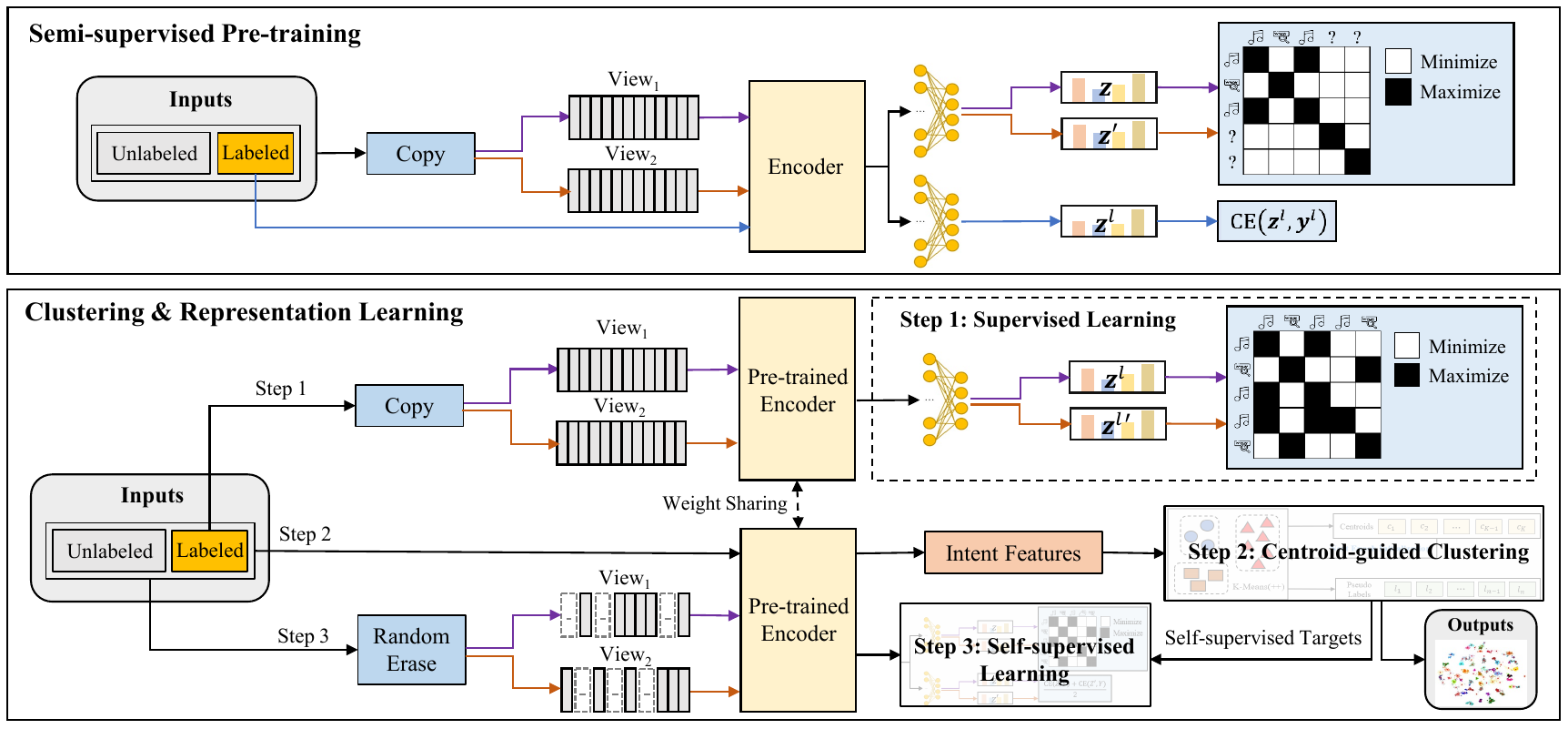}
    \caption{\label{semi-example} The pipeline of semi-supervised new intent discovery. On the basis of unsupervised new intent discovery, it enhances the pre-training stage by incorporating labeled data through the use of both semi-supervised contrastive and cross-entropy losses. It also improves the clustering and representation learning stage by adding a supervised contrastive learning step on labeled data to address the issue of \textit{catastrophic forgetting}.}
\end{figure*}

 In this section, we introduce a new clustering framework, USNID. The pipelines of unsupervised and semi-supervised new intent discovery are presented in Figure~\ref{unsup} and Figure~\ref{semi-example}. 
 
	\subsection{Intent Representation}
	\label{represent}
	The pre-trained BERT language model shows excellent performance in a wide range of NLP tasks~\cite{BERT}. Thus, it is adopted to extract deep intent representations. 
	
	Specifically, for each utterance $\boldsymbol{x}_{i}$, we take it as input to BERT in the needed format (i.e., the first token is [$\textrm{CLS}$]) and obtain its final hidden vectors $[C, T_1,..., T_L] \in \mathbb R^{(L+1) \times H}$ of each token  through non-linear projection layers of BERT, where $L$ is the length of the $i^{\textrm{th}}$ utterance, $H$ is the hidden size 768. The sentence  representation $\boldsymbol{s}_{i} \in \mathbb R^{H}$ is first obtained by applying $\operatorname{mean-pooling}$ operation on the hidden vectors of these tokens:
	\begin{align}
	\boldsymbol{s}_{i} = \operatorname{mean-pooling}([C, T_1,..., T_L]).
	\end{align} 
	
	Then, a fully-connected layer $h$ is added to enhance the capability to capture the complex semantics of high-dimensional text data, yielding the intent representation $\boldsymbol{I}_{i} \in \mathbb R^{D}$:
	\begin{align}
	\boldsymbol{I}_{i}=h(\boldsymbol{s}_i) = W_h\boldsymbol{s}_{i}+b_h,
	\end{align}
	where $D$ is the feature dimension,  $W_h \in \mathbb R^{H \times D}$ and $b_h \in \mathbb R^{D}$ are weight matrices and bias vectors, respectively.
	\subsection{Unsupervised New Intent Discovery}
	\subsubsection{Unsupervised Pre-training}
 \label{unsup-pretrain}
Initially, samples from different intent classes often overlap in the feature space, which can hinder the clustering optimization, leading to suboptimal clusters~\cite{zhang2021supporting}. Well-initialized intent representations, which should be distributed uniformly in the feature space and effectively reflect the data characteristics, can improve clustering performance and convergence~\cite{jain1999data,arthur2007k}.  Therefore, our goal in pre-training the model is to push apart distinct samples while capturing implicit semantic relationships between augmented pairs.

Given that only unlabeled data are available, a common way to construct positive pairs is to use two augmented views of the same sample, as suggested in~\cite{chen2020simple}. 
    In particular, let $\tilde{\boldsymbol{x}}_{i}$ and $\tilde{\boldsymbol{x}}_{i}'$ be two views of $\boldsymbol{x}_{i}$ in a mini-batch of $n$ samples. $\tilde{\boldsymbol{x}}_{i}$ and
    $\tilde{\boldsymbol{x}}_{i}'$ are treated as a positive pair, and they form negative pairs with the remaining $2n-2$ augmented samples. They are first encoded as intent representations $\tilde{\boldsymbol{I}}_{i}$ and $\tilde{\boldsymbol{I}}_{i}'$, as introduced in section~\ref{represent}. Then, we add a non-linear projection head $f_{1}^{u}:\mathbb R^{D}\rightarrow\mathbb R^{K}$ to obtain $\boldsymbol{z}_{i}$ and $\boldsymbol{z}_{i}'$. The unsupervised contrastive loss $\mathcal{L}_{\textrm{ucl}}$ is defined as:
    \begin{align}
	\mathcal{L}_{\textrm{ucl}}=-\frac{1}{2n}\sum_{i=1}^{2n} \log\frac{\exp(\textrm{sim}(\boldsymbol{z}_{i},\boldsymbol{z}_{i}')/\tau)}{\sum_{j=1}^{2n}\mathbb{I}_{[j\neq i]}\exp(\textrm{sim}(\boldsymbol{z}_{i},\boldsymbol{z}_{j})/\tau)},
	\end{align}
    where $\textrm{sim}(\boldsymbol{a},\boldsymbol{b})$ performs dot product on L2-normalized $\boldsymbol{a}$ and $\boldsymbol{b}$, $\tau$ is the temperature parameter, and $\mathbb{I}(\cdot)$ is an indicator function outputs 1 iff $j\neq i$ and 0 otherwise. 
   
    A simple yet effective method, \textit{random erase}, is used as a strong data augmentation for new intent discovery. Specifically, for an utterance $\boldsymbol{x}_{i}$ with length $L$, we randomly select $ \lfloor L \times a\%\rfloor$ different words and erase them from $\boldsymbol{x}_{i}$, where $a$ is the erase ratio in a sentence. The intuition is that this operation explicitly provides hard positive pairs (i.e., different missing sets of words) for contrastive learning, which is beneficial to capture the fine-grained semantic relations between different local word sets in a sentence. After pre-training, we remove the head used in contrastive learning to avoid any unwanted bias that might interfere with the subsequent steps. The rest of the backbone is saved for clustering and representation learning. 
	\subsubsection{Centroid-guided Clustering}
	\label{c-g-c}Partitioning clustering methods such as the K-Means algorithm can be used to discover intent-wise clusters.  However, its effectiveness can be compromised by the choice of initial centroids. In scenarios where the initial centroids are suboptimal, the algorithm risks falling into a local minima, resulting in unsatisfactory clustering. This shortcoming is addressed by K-Means++~\cite{arthur2007k}, which adopts a probabilistic approach to select new centroids, thereby improving convergence and achieving an optimal solution more quickly than the standard K-Means. Consequently, our work utilizes K-Means++ for clustering purposes.  
    
    

    We found that directly using K-Means++ still performs poorly due to a lack of guidance to help enhance the intent representation capability. Thus, we aim to use the clustering information to construct high-quality self-supervised signals for learning high-level intent representations. A natural way to do this is to use the cluster assignments as pseudo-labels for supervision. This is based on the typical clustering approach proposed by~\cite{caron2018deep}, where they alternate between clustering and optimizing the convnets based on predicting the cluster assignments. They demonstrate structured outputs of neural networks used as weakly supervised signals can also benefit the unsupervised representation learning. However, a challenge arises as the same sample could be assigned to different clusters across multiple iterations due to centroid selection randomness. Although Caron et al.~\cite{caron2018deep} propose randomly re-initializing classifier parameters before each training iteration to address this, this strategy fails to make effective use of historical training information\cite{zhan2020online}.
    
    
    

    In this work, we introduce a novel centroid-guided mechanism to address inconsistencies in self-supervised targets between training iterations and enhance knowledge retention in the classifier. Noting that while cluster assignments may fluctuate between iterations, cluster centroids remain relatively stable due to their global optimization as averaged features, we propose using these centroids as guidance. This approach aims to provide consistent self-supervised targets across training iterations and preserve the well-trained knowledge of the classifier, enhancing the overall effectiveness of the iterative process.
    
    In particular, the cluster centroids and assignments in the last and current training  iterations are denoted as $\boldsymbol{C}^{(t-1)}$, $\boldsymbol{y}^{(t-1)}$, and $\boldsymbol{C}^{(t)}$, $\boldsymbol{y}^{(t)}$, respectively. After the $(t-1)^{\textrm{th}}$ clustering,  $\boldsymbol{y}^{(t-1)}$ is used as supervision for feature learning, which helps capture similarity relationships of the samples close to $\boldsymbol{C}^{(t-1)}$. The updated representations are then used for the $(t)^{\textrm{th}}$ clustering, which generates $\boldsymbol{C}^{(t)}$. The intuition is that $\boldsymbol{C}^{(t)}$ and $\boldsymbol{C}^{(t-1)}$ have relatively consistent  distributions in the feature space, and $\boldsymbol{C}^{(t)}$ is aligned with $\boldsymbol{C}^{(t-1)}$ to obtain the optimal mapping $G_{\textrm{opt}}$ as below:
    \begin{align}
        G_{\textrm{opt}} = \underset{G}{\operatorname{argmin}}\left\{\sum_{i=1}^{K}\|\boldsymbol{C}_{i}^{(t)}-\boldsymbol{C}_{g_{i}}^{(t-1)}\|_{2}\right\},
        \label{align}
    \end{align}
    where $G: \{1,...,K\} \rightarrow \{1,...,K\}$ is a one-to-one mapping, $g_{i}=G(i)$ is the centroid index corresponding to $i$ in the last iteration. It can be optimized with the Hungarian algorithm~\cite{kuhn1955hungarian} to obtain $G_{\textrm{opt}}$. Then, $\boldsymbol{C}^{(t)}$ and $\boldsymbol{y}^{(t)}$ are updated by:
    \begin{align}
        \boldsymbol{C}_{i}^{(t)}&=\boldsymbol{C}_{g_{i}'}^{(t-1)},  \text { s.t. } \ g_{i}'=G_{\textrm{opt}}^{-1}(i), \forall i \in \{1,...,K\}, \\ y_{i}^{(t)}&=G_{\textrm{opt}}^{-1}(y_{i}^{(t-1)}),  \forall i \in \{1,...,N\},
    \end{align}
    where $G_{\textrm{opt}}^{-1}$ is the inverse mapping of $G_{\textrm{opt}}$. The preliminary results of this centroid-guided alignment strategy have been presented in our previous work~\cite{Zhang_Xu_Lin_Lyu_2021} and show substantial improvements compared with the re-initialization strategy. 
    
    However, this strategy is not efficient due to the high time cost of multiple clustering. The reason is that each clustering (i.e., K-Means++) also needs to initialize the first centroid at random, which may select sub-optimal centroids and lead to a degradation of convergence. Thus, finding the optimal solution will take a lot more time. To solve this problem, we propose a concise centroid-guided initialization strategy, aiming to leverage the historical clustering information to improve convergence. Specifically, K-Means++ is only performed at the first training iteration. Then, the cluster centroids produced in the $(t-1)^{\textrm{th}}$ training iteration are used to initialize K-Means, yielding $\boldsymbol{y}^{(t)}$ and $\boldsymbol{C}^{(t)}$:
    \begin{align}
           \boldsymbol{y}^{(t)}, \boldsymbol{C}^{(t)}&=\begin{cases}
        \operatorname{\textrm{K-Means++}}\left(\boldsymbol{I}^{(t)}\right), & \text { if } t=0, \nonumber \\ 
        \operatorname{\textrm{K-Means}}\left(\boldsymbol{I}^{(t)}, \boldsymbol{C}^{(t-1)}\right), & \text { if } t \geq 1.
        \end{cases}\\ 
        \text { s.t. } \boldsymbol{I}^{(t)}&=\textrm{Learn}(\boldsymbol{I}^{(t-1)}, \boldsymbol{y}^{(t-1)}), t\geq1; \ t\in \mathbb{N},
    \end{align} 
    where $\boldsymbol{I}^{(0)}$ denotes the initial intent representations after pre-training, $\textrm{Learn}(\boldsymbol{I}, \boldsymbol{y})$ denotes the representation learning process with pseudo-labels $\boldsymbol{y}$ as supervision, which will be introduced in section~\ref{self-sup} in detail. With a random centroid initialization, we need to perform alignment between $\boldsymbol{C}^{(t-1)}$ and $\boldsymbol{C}^{(t)}$ to obtain $G_{\textrm{opt}}$ as in Eq.~\ref{align}. Interestingly, we found that this strategy does not need the alignment process, as experiments show that $G_{\textrm{opt}}(i)=i, \ \forall i \in \{1,...,K\}$ usually works, and $\boldsymbol{y}^{(t)}$ can be directly used as aligned targets, which also converge well. It is reasonable because 
   centroid initialization ensures the stability of cluster allocation targets between adjacent clustering and is beneficial to find the optimal solution with prior knowledge of previous cluster centroids.

The stopping criterion of training is to compare cluster assignments between adjacent clustering $\boldsymbol{y}^{(t)}$ and $\boldsymbol{y}^{(t-1)}$:
   \begin{align}
       \delta =  \frac{\sum_{i=1}^N \mathbb{I} \left\{y^{(t)}_i\neq y^{(t-1)}_i\right\}}{N},
   \end{align}
      where $\mathbb{I}(\cdot)$ is the indicator function that outputs 1 only if the condition holds. Otherwise, it outputs 0. $\delta$ indicates the proportion of the difference between $\boldsymbol{y}^{(t)}$ and $\boldsymbol{y}^{(t-1)}$, which values are in the range of [0, 1]. It can well reflect the convergence of the proposed clustering algorithm. The procedure will be stopped when $\delta$ is smaller than some threshold $\delta_{\textrm{th}}$. During the inference phase, we perform another K-Means using the cluster centroids that have been well-trained in the previous clustering as the initialization. 
  
    \subsubsection{Self-supervised Learning}
    \label{self-sup}
    After each clustering, we learn discriminative intent representations to further promote the subsequent clustering. We achieve this through a dual learning strategy that operates at both the instance and cluster levels, drawing inspiration from the methodologies presented in recent studies~\cite{van2020scan,caron2020unsupervised}. Instance-level learning focuses on ensuring that similar instances are allocated to the same class while differentiating them from instances that belong to other classes. This proves especially effective under strong data augmentations~\cite{li2022twin}. It fosters the development of intra-class compactness and inter-class separability within our model, both of which are fundamental properties that aid in clustering. 

    Simultaneously, we perform cluster-level learning by updating the model parameters based on the aligned cluster assignments $\boldsymbol{y}^{a}$. This approach enhances the model's discriminative power to discern and classify intents by predicting cluster assignments. By concurrently learning at both the instance and cluster levels, we gain a more comprehensive and nuanced understanding of the data. This strategy, which focuses on the fine-grained details of individual instances and the overarching characteristics of clusters, significantly enhances the overall performance and effectiveness of our clustering efforts.


    
    
    
    
    In particular, we first perform \textit{random erase} data augmentation and use the pre-trained encoder in section~\ref{unsup-pretrain} to extract two views of intent representations $\tilde{\boldsymbol{I}}$ and
    $\tilde{\boldsymbol{I}}'$. To capture cluster-level information, we perform the classification loss $\mathcal{L}_{\textrm{cls}}$:
    \begin{align}
        \mathcal{L}_{\textrm{cls}}&=\frac{\mathcal{L}_{\textrm{ce}}(\boldsymbol{z},\boldsymbol{y}^{a})+\mathcal{L}_{\textrm{ce}}(\boldsymbol{z}',\boldsymbol{y}^{a})}{2},\\
        \mathcal{L}_{\textrm{ce}}(\boldsymbol{z},\boldsymbol{y}^{a})&=-\frac{1}{N}\sum_{i=1}^{N} \log\frac{\exp({(\boldsymbol{z}_{i})}^{y_{i}^{a}})}{\sum_{j=1}^{K}\exp({(\boldsymbol{z}_{i})}^{j})},\label{ce}
    \end{align}    
    where  $\boldsymbol{z}$ and $\boldsymbol{z}'$ are obtained with $f_{2}^{u}:\mathbb R^{D}\rightarrow\mathbb R^{K}$, and $\mathcal{L}_{\textrm{ce}}$ is the cross-entropy loss.  As $\boldsymbol{y}^{a}$ is relatively consistent between adjacent clustering, it can provide stable supervised signals for learning cluster-level patterns. Two augmented views are used as hard examples for classification, which is beneficial to enhance the model's discrimination ability. 
 
    To capture instance-level information, we aim to pull samples from the same class close to each other and push samples from different classes away from each other. For this purpose, we apply the supervised contrastive loss as suggested in~\cite{khosla2020supervised}:
     \begin{align}
     \label{scl}
        &\mathcal{L}_{\textrm{scl}}=\nonumber \\ 
        &-\frac{1}{2n}\sum_{i=1}^{2n}\frac{1}{|{P}(i)|}\sum_{p\in\mathcal{P}(i)}\log\frac{\exp(\textrm{sim}(\boldsymbol{z}_{i},\boldsymbol{z}_{p})/\tau)}{\sum_{j=1}^{2n}\mathbb{I}_{[j\neq i]}\exp(\textrm{sim}(\boldsymbol{z}_{i},\boldsymbol{z}_{j})/\tau)},
    \end{align}
    where $\boldsymbol{z}_{i}$ is obtained with $f_{3}^{u}:\mathbb R^{D}\rightarrow\mathbb R^{K}$, $n$ is the number of samples in a mini-batch, $\mathcal{P}(i)$ is the set of indices of augmented samples with the same class of $\boldsymbol{z}_{i}$, and $|\cdot|$ denotes the size of a set. 

    The overall loss of self-supervised learning  can be written as follows:
    \begin{align}
    \mathcal{L}_{\textrm{self-sup}}=\mathcal{L}_{\textrm{cls}}+\mathcal{L}_{\textrm{scl}}.
    \end{align}
    That is, we joint train $\mathcal{L}_{\textrm{cls}}$ and $\mathcal{L}_{\textrm{scl}}$ to learn both cluster-level and instance-level characteristics, which is helpful to obtain friendly representations for clustering. 

    While some SOTA unsupervised clustering methods, such as CC and SCCL, also perform representation learning at both the instance and cluster levels, our method stands apart due to three key differentiating factors. First, our method integrates a pre-training strategy to discern distinct instances during the initial stages while concurrently learning latent correlations. This serves as an effective regularization procedure, facilitating more amenable representations for subsequent clustering. Second, instead of relying solely on weak pairwise constraints employed by CC and SCCL, our method generates specific pseudo-labels as self-supervised signals for each sample. Importantly, the cluster-level learning objectives in our model can explicitly differentiate between various intent classes. Third, our approach introduces an innovative perspective on the creation of high-quality self-supervised targets. Particularly, the centroid-guided mechanism enables effectively leveraging historical clustering data to generate aligned instance-level pseudo-labels. This not only greatly enhances clustering performance but also leads to excellent convergence. These distinct features enable our method to outperform current SOTA approaches, achieving substantial improvements of over 30\% in standard clustering metrics.
	\subsection{Semi-supervised New Intent Discovery}
	\subsubsection{Semi-supervised Pre-training}
 In the pre-training phase, we hope to fully utilize the limited annotated intent data to provide well-initialized representations for clustering. For this purpose, we perform data augmentation on both labeled and unlabeled data and mix them in nearly equal ratios within a mini-batch for contrastive learning. The positive pairs include: (a) samples with the same class in the labeled data, (b) each sample with its augmented view in both labeled and unlabeled data. Thus, we propose the semi-supervised contrastive loss:
     \begin{align}
        &\mathcal{L}_{\textrm{semi-scl}}=\nonumber \\ 
        &-\frac{1}{2n}[\sum_{{z}_{i}\in\{\boldsymbol{z}^{l}\}}\frac{1}{|\mathcal{P}(i)|}\sum_{p\in\mathcal{P}(i)}\log\frac{\exp(\textrm{sim}(\boldsymbol{z}_{i},\boldsymbol{z}_{p})/\tau)}{\sum_{j=1}^{2n}\mathbb{I}_{[j\neq i]}\exp(\textrm{sim}(\boldsymbol{z}_{i},\boldsymbol{z}_{j})/\tau)}\nonumber \\
        &+\sum_{{z}_{i}\in\{\boldsymbol{z}^{u}\}}\log\frac{\exp(\textrm{sim}(\boldsymbol{z}_{i},\boldsymbol{z}_{i}')/\tau)}{\sum_{j=1}^{2n}\mathbb{I}_{[j\neq i]}\exp(\textrm{sim}(\boldsymbol{z}_{i},\boldsymbol{z}_{j})/\tau)}],
    \end{align}
where $\boldsymbol{z}_{i}$ and $\boldsymbol{z}_{i}'$ are obtained with $f_{1}^{s}:\mathbb R^{D}\rightarrow\mathbb R^{K^{\textrm{known}}}$. $\{\boldsymbol{z}^{l}\}$ and $\{\boldsymbol{z}^{u}\}$ are the sets of labeled and unlabeled data, respectively, which satisfy $|\{\boldsymbol{z}^{l}\}|+|\{\boldsymbol{z}^{u}\}|=2n$. Here we use a simple data augmentation method, \textit{dropout}~\cite{gao2021simcse}, which is efficient and works well. It uses dropout masks in transformers to produce positive pairs with the same sample feed-forward twice in neural networks. Moreover, we add a cross-entropy loss $\mathcal{L}_{\textrm{ce}}$ with supervised signals of labeled data $\boldsymbol{y}^{l}$ to enhance the  discrimination ability for known classes. The final loss of semi-supervised pre-training is defined as:
    \begin{align}
    \mathcal{L}_{\textrm{semi-pre}}=\mathcal{L}_{\textrm{semi-scl}}+\mathcal{L}_{\textrm{ce}}(\boldsymbol{z}^{l},\boldsymbol{y}^{l}),
    \end{align}
where $\boldsymbol{z}^{l}$ is obtained with $f_{2}^{s}:\mathbb R^{D}\rightarrow\mathbb R^{K^{\textrm{known}}}$. Similar to unsupervised pre-training, we use the pre-trained network without $f_{1}^{s}$ and $f_{2}^{s}$ in the subsequent steps.
\subsubsection{Clustering and Representation Learning}
After pre-training, we can directly use all data to perform clustering and representation learning as in unsupervised new intent discovery.  However, as the training iteration goes on, we notice that some labeled samples from the same class may be assigned to different clusters when mixed with unlabeled data for clustering, which is also described as the \textit{catastrophic forgetting} phenomenon in~\cite{han2019automatically}. To alleviate this problem, we propose using supervised contrastive learning with labeled data at the beginning of each training iteration. Specifically, we use the \textit{dropout} strategy to generate augmented samples and add a new head $f_{3}^{s}:\mathbb R^{D}\rightarrow\mathbb R^{K}$ to perform $\mathcal{L}_{\textrm{scl}}$ as in Eq.~\ref{scl}. It can not only strengthen the \textit{memory} of supervised similarity relationships but also be beneficial to guide the subsequent clustering process. 

Then, we successively carry out centroid-guided clustering (section~\ref{c-g-c}) and self-supervised learning (section~\ref{self-sup})  in the rest of each training iteration. $f_{3}^{s}$ and another head $f_{4}^{s}:\mathbb R^{D}\rightarrow\mathbb R^{K}$ are used for learning instance-level and cluster-level information, respectively. 
	
\begin{table*}[t!]\tiny
    \caption{ \label{datasets}  Statistics of BANKING, CLINC150, and StackOverflow datasets. \# indicates the total number of sentences. The unsupervised setting only contains new intents. In the semi-supervised setting, we randomly select 25\%, 50\%, and 75\% intents as known and treat the remaining as new intents.}
    \centering
    \resizebox{2\columnwidth}{!}{
    \begin{tabular}{@{} llccccc @{}}
        \toprule
        Dataset & \#Known Classes + \#New Classes & \#Training & \#Validation & \#Testing & Vocabulary & Length (max / mean) \\
        \midrule
        BANKING & 0 + 77 / 19 + 58 / 39 + 38 / 58 + 19 & 9,003 & 1,000 & 3,080 & 5,028 & 79 / 11.91 \\ 
        CLINC150 & 0 + 150 / 38 + 112 / 75 + 75 / 113 + 37 & 18,000 & 2,250 & 2,250 & 7,283 & 28 / 8.31 \\
        StackOverflow & 0 + 20 / 5 + 15 / 10 + 10 / 15 + 5 & 12,000 & 2,000 & 6,000 & 17,182 & 41 / 9.18 \\
        \bottomrule
    \end{tabular}}
\end{table*}
 
\subsection{Estimate the Cluster Number $K$}
\label{est-k}
To deal with an unknown cluster number $K$, we propose a simple yet effective method for estimating $K$ in unsupervised and semi-supervised new intent discovery. Specifically, the intent representations $\boldsymbol{I}$ after pre-training are first used to perform K-Means++ with a large assigned cluster number K'. Though the pre-training phase lacks explicit supervised signals for distinguishing fine-grained clusters, it can still help capture weak semantic similarity relations using positive augmented samples or limited labeled data. 

The assumption is that real clusters tend to be confident of having much more samples that are similar to each other. Specifically, in the unsupervised setting, we remove the low-confidence clusters and estimate $K$ by:
\begin{align}
    K=\sum_{k=1}^{K'} \mathbb{I} \left\{|\mathcal{C}_k|\geq t\right\},
\label{un_k}
\end{align}
where $\mathcal{C}_k=\{\boldsymbol{x}_{i}|y_{i}=k, i=1,2,...,N\}$, $t$ is a threshold defined as the mean cluster size $\frac{\sum_{k=1}^{K'}|\mathcal{C}_k|}{K'}$. 

In the semi-supervised setting, we have access to a set of known intent classes with a number $K^{\textrm{known}}$ through limited labeled data. The goal is to estimate the number of new intent classes $K^{\textrm{new}}$. Since the unlabeled data come from both known and new classes, we need to first distinguish the known intent clusters from clustering results. For this purpose, we propose to use the limited labeled data as prior knowledge for cluster induction. In particular, we perform the Hungarian algorithm to obtain the alignment projection $G_{\textrm{opt}}'$ between the labeled centroids $\boldsymbol{C}^l \in \mathbb R^{K^{\textrm{known}} \times D}$ and the cluster centroids $\boldsymbol{C} \in \mathbb R^{K' \times D}$ in the Euclidean space:
\begin{align}
    G_{\textrm{opt}}' = \underset{G'}{\operatorname{argmin}}\left\{\sum_{i=1}^{K^{\textrm{known}}}\|\boldsymbol{C}_{i}^{l}-\boldsymbol{C}_{b_{i}}\|_{2}\right\},
    \label{align}
\end{align}
where $G': \{1,...,K^{\textrm{known}}\} \rightarrow \{1,...,K'\}$, $b_{i}=G'(i)$ and $i$ are the corresponding centroid indices, and $\boldsymbol{C}^l$ is calculated by averaging intent representations of each class in the labeled samples. Then, we can find the set of known intent cluster indices $S=\{G_{opt}^{'}(i)\}_{i=1}^{K^{\textrm{known}}}$, and $K^{\textrm{new}}$ is calculated by:
\begin{align}
K^{\textrm{new}}=\sum_{k=1}^{K'} \mathbb{I} \left\{(|\mathcal{C}_k|\geq t) \wedge (k\notin S)\right\},
\end{align}
where $t$ is the same as in the unsupervised setting. The total cluster number $K$ is the summation of $K^{\textrm{known}}$ and $K^{\textrm{new}}$.
\section{Experiments}
\subsection{Datasets}
We evaluate the new intent discovery performance with three challenging benchmark datasets: BANKING~\cite{Casanueva2020}, CLINC150~\cite{larson-etal-2019-evaluation}, and StackOverflow~\cite{xu-etal-2015-short}. The detailed statistics for these datasets are shown in Table~\ref{datasets}. 

The BANKING dataset is a collection of customer service queries specifically from the banking domain, comprising 13,083 queries across 77 classes. We follow the data splits in~\cite{Casanueva2020} and create a validation set of 1,000 randomly sampled utterances from the original training set.

The CLINC150 dataset is an out-of-scope intent classification dataset with 150 classes across ten domains. Since the out-of-scope utterances lack specific intent annotations for evaluation, we only use the 22,500 in-scope queries in this work. The dataset is split into training, validation, and testing sets by 8:1:1.

The StackOverflow dataset originally contains 3,370,528 technical question titles on Kaggle.com\footnote{https://www.kaggle.com/competitions/predict-closed-questions-on-stack-overflow/data}. In this work, we use the curated version of the dataset presented in~\cite{xu-etal-2015-short}, consisting of 20,000 samples across 20 classes. The dataset is split into training, validation, and testing sets by 6:1:3. 

\begin{table*}[t!]\tiny
    \caption{ \label{results-main}  
        Results of new intent discovery on the three datasets. KCR denotes the known class ratio, with 0\% for unsupervised and 25\%, 50\%, and 75\% for semi-supervised settings. The proposed method USNID is significantly better than others with  $p$-value $<$ 0.05 ($\dagger$) and $p$-value $<$ 0.1 (*) using t-test.}
    \centering
    \resizebox{2\columnwidth}{!}{
    \begin{tabular}{@{\extracolsep{15pt}}c|lcccccccccc}
        \toprule
        \multirow{2}{*}{KCR} 
        & \multirow{2}{*}{Methods} & \multicolumn{3}{c}{BANKING} & \multicolumn{3}{c}{CLINC150} & \multicolumn{3}{c}{StackOverflow} \\ \cline{3-5}\cline{6-8}\cline{9-11}  \addlinespace[0.1cm]
        & & NMI & ARI & ACC & NMI & ARI & ACC & NMI & ARI & ACC  \\ 
        \hline\addlinespace[0.1cm]
        \multirow{8}{*}{0\%} 
         &KM  &49.30$\dagger$&13.04$\dagger$&28.62$\dagger$&71.05$\dagger$&27.72$\dagger$&45.76$\dagger$&19.87$\dagger$&5.23$\dagger$&23.72$\dagger$\\
         &AG   &53.28$\dagger$&14.64$\dagger$&31.62$\dagger$&72.21$\dagger$&27.05$\dagger$&44.13$\dagger$&25.54$\dagger$&7.12$\dagger$&28.50$\dagger$\\
     &SAE-KM  &59.80$\dagger$&23.59$\dagger$&37.07$\dagger$&73.77$\dagger$&31.58$\dagger$&47.15$\dagger$&44.96$\dagger$&28.23$\dagger$&49.11$\dagger$\\
         &DEC  &62.65$\dagger$&25.32$\dagger$&38.60$\dagger$&74.83$\dagger$&31.71$\dagger$&48.77$\dagger$&58.76$\dagger$&36.23$\dagger$&59.49$\dagger$\\
         &DCN  &62.72$\dagger$&25.36$\dagger$&38.59$\dagger$&74.77$\dagger$&31.68$\dagger$&48.69$\dagger$&58.75$\dagger$&36.23$\dagger$&59.48$\dagger$\\
         &CC &44.89$\dagger$&9.75$\dagger$&21.51$\dagger$&65.79$\dagger$&18.00$\dagger$&32.69$\dagger$&19.06$\dagger$&8.79$\dagger$&21.01$\dagger$\\
         &SCCL &63.89$\dagger$&26.98$\dagger$&40.54$\dagger$&79.35$\dagger$&38.14$\dagger$&50.44$\dagger$&69.11$\dagger$&34.81$\dagger$&68.15\\
         &USNID & \textbf{75.30} & \textbf{43.33} & \textbf{54.83} & \textbf{91.00} & \textbf{68.54} & \textbf{75.87} & \textbf{72.00} & \textbf{52.25} & \textbf{69.28}\\
         \midrule
         \midrule
        \multirow{9}{*}{25\%} 
        & KCL&52.70$\dagger$&18.58$\dagger$&26.03$\dagger$&67.98$\dagger$&24.30$\dagger$&29.40$\dagger$&30.42$\dagger$&17.66$\dagger$&30.69$\dagger$ \\
        & MCL&47.88$\dagger$&14.43$\dagger$&23.29$\dagger$&62.76$\dagger$&18.21$\dagger$&28.52$\dagger$&26.68$\dagger$&17.54$\dagger$&31.46$\dagger$ \\
        & DTC&55.59$\dagger$&19.09$\dagger$&31.75$\dagger$&79.35$\dagger$&41.92$\dagger$&56.90$\dagger$&29.96$\dagger$&17.51$\dagger$&29.54$\dagger$\\
        & GCD&59.74$\dagger$&26.04$\dagger$&38.50$\dagger$&83.70$\dagger$&52.23$\dagger$&64.82$\dagger$&29.69$\dagger$&15.48$\dagger$&34.84$\dagger$\\
        & CDAC+&66.39$\dagger$&33.74$\dagger$&48.00$\dagger$&84.68$\dagger$&50.02$\dagger$&66.24$\dagger$&46.16$\dagger$&30.99$\dagger$&51.61$\dagger$\\
        & DeepAligned&70.50$\dagger$&37.62$\dagger$&49.08$\dagger$&88.97$\dagger$&64.63$\dagger$&74.07$\dagger$&50.86$\dagger$&37.96$\dagger$&54.50$\dagger$\\
        & DCSC&78.18&49.75&60.15&91.70&72.68&79.89&-&-&-\\
        & MTP-CLNN&80.04$\dagger$&52.91$\dagger$&65.06&93.17$\dagger$&76.20$\dagger$&\textbf{83.26}&73.35&54.80$\dagger$&74.70\\
        & USNID&\textbf{81.94}&\textbf{56.53}&\textbf{65.85}&\textbf{94.17}&\textbf{77.95}&83.12&\textbf{74.91}&\textbf{65.45}&\textbf{75.76}\\
        \midrule
        \midrule
        \multirow{9}{*}{50\%} 
        &KCL&63.50$\dagger$&30.36$\dagger$&40.04$\dagger$&74.74$\dagger$&35.28$\dagger$&45.69$\dagger$&53.39$\dagger$&41.74$\dagger$&56.80$\dagger$\\
        &MCL&62.71$\dagger$&29.91$\dagger$&41.94$\dagger$&76.94$\dagger$&39.74$\dagger$&49.44$\dagger$&45.17$\dagger$&36.28$\dagger$&52.53$\dagger$\\
        &DTC&69.46$\dagger$&37.05$\dagger$&49.85$\dagger$&83.01$\dagger$&50.44$\dagger$&64.39$\dagger$&49.80$\dagger$&37.38$\dagger$&52.92$\dagger$\\
        &GCD&66.97$\dagger$&35.07$\dagger$&48.35$\dagger$&87.12$\dagger$&59.86$\dagger$&70.89$\dagger$&50.60$\dagger$&31.98$\dagger$&55.27$\dagger$\\
        &CDAC+&67.30$\dagger$&34.97$\dagger$&48.55$\dagger$&86.00$\dagger$&54.87$\dagger$&68.01$\dagger$&46.21$\dagger$&30.88$\dagger$&51.79$\dagger$\\
        &DeepAligned&76.67$\dagger$&47.95$\dagger$&59.38$\dagger$&91.59$\dagger$&72.56$\dagger$&80.70$\dagger$&68.28$\dagger$&57.62$\dagger$&74.52$\dagger$\\
        &DCSC&81.19&56.94&68.30&93.75&78.82&84.57& - & - & -\\
        &MTP-CLNN&83.42$\dagger$&60.17$\dagger$&70.97*&94.30$\dagger$&80.17$\dagger$&86.18&76.66$\dagger$&62.24$\dagger$&80.36\\
        & USNID & \textbf{85.05} & \textbf{63.77} & \textbf{73.27}&\textbf{95.45}&\textbf{82.87}&\textbf{87.22}&\textbf{78.77}&\textbf{71.63}&\textbf{82.06}\\
        \midrule
        \midrule
        \multirow{9}{*}{75\%} 
        & KCL&72.75$\dagger$&45.21$\dagger$&59.12$\dagger$&86.00$\dagger$&58.62$\dagger$&68.89$\dagger$&63.98$\dagger$&54.28$\dagger$&68.69$\dagger$\\
        & MCL&74.42$\dagger$&48.06$\dagger$&61.56$\dagger$&87.26$\dagger$&61.21$\dagger$&70.27$\dagger$&63.44$\dagger$&56.11$\dagger$&71.71$\dagger$\\
        & DTC&74.44$\dagger$&44.68$\dagger$&57.16$\dagger$&89.19$\dagger$&67.15$\dagger$&77.65$\dagger$&63.05$\dagger$&53.83$\dagger$&71.04$\dagger$\\
        & GCD&72.48$\dagger$&43.36$\dagger$&57.32$\dagger$&89.42$\dagger$&65.98$\dagger$&76.78$\dagger$&61.99$\dagger$&43.61$\dagger$&66.73$\dagger$\\
        & CDAC+&69.54$\dagger$&37.78$\dagger$&51.07$\dagger$&85.96$\dagger$&55.17$\dagger$&67.77$\dagger$&58.23$\dagger$&40.95$\dagger$&64.57$\dagger$\\
        & DeepAligned&79.39$\dagger$&53.09$\dagger$&64.63$\dagger$&93.92$\dagger$&79.94$\dagger$&86.79$\dagger$&73.28$\dagger$&60.09$\dagger$&77.97$\dagger$\\
        & DCSC& 84.65 & 64.55 & 75.18 & 95.28 & 84.41 & 89.70 & - & - & -\\
        & MTP-CLNN &86.19$\dagger$&66.98$\dagger$&77.22&95.45$\dagger$&84.30$\dagger$&89.46*&77.12$\dagger$&69.36$\dagger$&82.90$\dagger$\\
        & USNID & \textbf{87.41} & \textbf{69.54} & \textbf{78.36}&\textbf{96.42}&\textbf{86.77}&\textbf{90.36}&\textbf{80.13}
        &\textbf{74.90}&\textbf{85.66}\\
         \bottomrule 
    \end{tabular}}
\end{table*}
\subsection{Baselines}
\subsubsection{Unsupervised Clustering}
The unsupervised clustering baselines include traditional machine learning methods: KM~\cite{macqueen1967some}, AG~\cite{gowda1978agglomerative}, SAE-KM, and deep clustering methods: DEC~\cite{xie2016unsupervised}, DCN~\cite{yang2017towards}, CC~\cite{li2021contrastive}, SCCL~\cite{zhang2021supporting}.

For KM and AG, the intent representations are extracted with GloVe~\cite{pennington-etal-2014-glove} by averaging the pre-trained 300-dimensional token embeddings in the sentence. For SAE-KM, DEC, and DCN, a stacked autoencoder (SAE)~\cite{vincent2010stacked} is used to capture semantically meaningful and discriminative representations~\cite{xie2016unsupervised}. Since CC is a method in the field of CV, we adapt it to this task by using BERT to extract intent representations. For SCCL, we use the Sentence transformer~\cite{reimers-gurevych-2019-sentence} as the backbone suggested in~\cite{zhang2021supporting}.

\subsubsection{Semi-supervised Clustering}
The semi-supervised clustering baselines contain a series of SOTA methods in related fields, including: constrained clustering: KCL~\cite{hsu2018learning}, MCL~\cite{hsu2018multiclass}, novel class discovery: DTC~\cite{Han2019learning}, GCD~\cite{vaze2022generalized}, and new intent discovery: CDAC+~\cite{lin2020discovering}, DeepAligned~\cite{Zhang_Xu_Lin_Lyu_2021},  DCSC~\cite{wei2022semi}, MTP-CLNN~\cite{zhang2022new}. 

Since KCL, MCL, DTC, and GCD are used for CV tasks, we adapt them to our task by using the BERT backbone. For MTP-CLNN, the parameter of top-K nearest neighbors is set to 50, 60, 300 for BANKING, CLINC150, and StackOverflow, respectively, which is used or calculated as in~\cite{zhang2022new}. For a fair comparison, the external dataset is not used in MTP-CLNN as other baselines. 
	
\subsection{Evaluation Metrics}
Three widely used metrics are adopted to evaluate the clustering performance, including normalized mutual information (NMI), adjusted rand index (ARI), and accuracy (ACC). The higher values of these metrics indicate better performance. Specifically, NMI is defined as:
\begin{align}
    \textrm{NMI}(\mathbf{y}^{gt}, \mathbf{y}^{p}) &=  \frac{\it{MI}(\mathbf{y}^{gt}, \mathbf{y}^{p})}{\frac{1}{2} (H(\mathbf{y}^{gt}) + H(\mathbf{y}^{p}))},
\end{align}
where $\mathbf{y}^{gt}$ and $\mathbf{y}^{p}$  are the ground-truth  and predicted labels, respectively. $\it{MI}(\mathbf{y}^{gt}, \mathbf{y}^{p})$ represents the mutual information between $\mathbf{y}^{gt}$ and $\mathbf{y}^{p}$, and $H(\cdot)$ is the entropy.  $\it{MI}(\mathbf{y}^{gt}, \mathbf{y}^{p})$ is normalized by the arithmetic mean of $H(\mathbf{y}^{gt})$ and $H(\mathbf{y}^{p})$, and the values of $\textrm{NMI}$ are in the range of [0, 1]. 

ARI is defined as:
\begin{align}
    \textrm{ARI} &= \frac{
    \sum_{i, j}\binom{n_{i, j}}{2}-[\sum_{i}\binom{u_{i}}{2}\sum_{j}\binom{v_{j}}{2}] / \binom{n}{2}
    }
    {
    \frac{1}{2}[\sum_{i}\binom{u_{i}}{2}+\sum_{j}\binom{v_{j}}{2}]-[\sum_{i}\binom{u_{i}}{2}\sum_{j}\binom{v_{j}}{2}]/\binom{n}{2}
    },
\end{align}
where $u_{i}=\sum_{j}n_{i,j}$, and $v_{j}=\sum_{i}n_{i,j}$. $n$ is the number of samples, and $n_{i,j}$ is the number of the samples that have both the $i^{\textrm{th}}$ predicted label  and the $j^{\textrm{th}}$ ground-truth label. The values of ARI are in the range of [-1, 1]. 

ACC is defined as:
\begin{align}
     \textrm{ACC}(\mathbf{y}^{gt}, \mathbf{y}^{p}) &=\max _m \frac{\sum_{i=1}^n \mathbb{I} \left\{y^{gt}_i=m\left(y^{p}_i\right)\right\}}{n},
\end{align}
where $m$ is a one-to-one mapping between the ground-truth label $\mathbf{y}^{gt}$ and predicted label $\mathbf{y}^{p}$ of the $i^{\textrm{th}}$ sample. The Hungarian algorithm is used to obtain the best mapping $m$ efficiently. The values of ACC are in the range of [0, 1].

\subsection{Experimental Settings}
In the unsupervised setting, we use the data in both training and validation sets for unsupervised learning with the aim of discovering intent-wise clusters in the testing set. In the semi-supervised setting, we randomly select a certain percentage (25\%, 50\%, and 75\%) of known intent classes. In the training set, we keep labels for a limited portion (10\%) of the data from these known classes, while the remaining data from known classes and all data from new classes are unlabeled. To simulate  real-world scenarios, the validation set only contains labeled data from known classes. The goal is to find known and discover new intent-wise clusters in the testing set.

We use the pre-trained BERT language model with 12 transformer layers as the backbone, which is implemented in~\cite{wolf2020transformers}.  For all experiments, we use AdamW~\cite{loshchilov2018decoupled} as the optimizer to train the model. The training process consists of 100 epochs, with a batch size of 128, and learning rates searched from \{1e-5, 2e-5, 5e-5\}. The intent feature dimension $D$ is 768.  All the non-linear projection heads $\{f_{i}^{u}\}_{i=1}^{3}$ and $\{f_{i}^{s}\}_{i=1}^{4}$ have the same architecture of $W\sigma(\cdot)+b$, where $W$ and $b$ are the weight matrix and bias term of a single linear layer, and $\sigma$ is the Tanh activation function. 

For the unsupervised setting, the temperature $\tau$ and \textit{random erase} ratio $a$ are set to 0.07 and 0.5, respectively. In addition, we fine-tune with the parameters of the last transformer layer as suggested in~\cite{lin2020discovering,Zhang_Xu_Lin_Lyu_2021}, which can improve  training efficiency and maintain good performance. For the semi-supervised setting, $\tau$ and $a$ are set to \{0.05, 0.4\} for BANKING and StackOverflow, and \{0.1, 0.3\} for CLINC150. The pre-training stage follows the same fine-tuning strategy as in the unsupervised setting, while the clustering and representation learning stage fine-tunes with all the transformer layers as in~\cite{zhang2022new}, which can fully explore high-level semantics with the guidance of labeled data. The K-Means++ clustering algorithm is implemented with the Scikit-learn~\cite{pedregosa2011scikit} toolkit. The threshold $\delta_{\textrm{th}}$ for stopping the training procedure is set to 0.0005. We implement our approach in PyTorch 1.8.1 and run experiments on NVIDIA Geforce RTX 3090 GPUs. For all experiments, we report the averaged results over ten runs with random seeds of 0-9. All the baselines are built upon our TEXTOIR platform~\cite{zhang-etal-2021-textoir}.

\begin{table*}[t!]\tiny
\caption{ \label{results-ablation}  
    Ablation studies of USNID. "w / o" means removing a component of USNID. Detailed information of each component can be seen in section~\ref{ablation}.
}
\renewcommand{\arraystretch}{1.2} 
\centering
\resizebox{2\columnwidth}{!}{
\begin{tabular}{@{\extracolsep{20pt}}c|llccccccccc}
    \toprule
    \centering
   
\multirow{2}{*}{KCR}	& \multirow{2}{*}{Stage 1}  & \multirow{2}{*}{Stage 2}  & \multicolumn{3}{c}{BANKING} & \multicolumn{3}{c}{CLINC150} & \multicolumn{3}{c}{StackOverflow}\\
    \addlinespace[0.05cm] \cline{4-6}\cline{7-9}\cline{10-12} \addlinespace[0.05cm]
 &	&  & NMI & ARI & ACC & NMI & ARI & ACC & NMI & ARI & ACC\\
    \midrule
     \specialrule{0em}{1pt}{1pt}
    \multirow{5}{*}[-1.5ex]{{0\%}} &
    w/o UCL & Full &66.69   &30.17   &41.65   &84.73   &50.76   &62.00   &36.61   &18.06   &31.89\\
      \cline{2-12}
    \specialrule{0em}{1pt}{1pt}
        &\multirow{4}{*}{Full} 
    & K-Means++ &62.16&27.97&40.69&77.56&38.36&53.26&22.08&9.05&23.66\\
    & & w/o CGM & 64.53&30.03
&42.17&82.68&48.42&60.57&20.20&6.66&19.94\\
    & &w/o CE &72.52&39.45&50.36&87.67&59.64&68.89&41.79&21.50&39.48\\
    & & Full  & \textbf{75.30} & \textbf{43.33} & \textbf{54.83} & \textbf{91.00} & \textbf{68.54} & \textbf{75.87}&
        \textbf{72.00} & \textbf{52.25} & \textbf{69.28}\\
    \midrule
    \midrule
     \specialrule{0em}{1pt}{1pt}
    \multirow{6}{*}[-1.5ex]{{25\%}} &
    w/o Semi-SCL & Full &80.06 &52.01  & 61.43   &92.99   &73.24   &78.85   &71.04   &60.73   &71.52\\
    \cline{2-12}
    \specialrule{0em}{1pt}{1pt}
        &\multirow{5}{*}{Full} 
    & K-Means++ &65.99&33.71&48.68&83.21&52.16&65.13&48.10&33.66&53.93\\
     & & w/o Self-Sup  &  71.39 & 41.00  &  55.12 &89.75   &67.41   &76.76  &57.44   &43.05   &63.85\\
    & & w/o CGM  &  72.25 & 42.32  & 55.54 &90.06  &67.66   &76.17  &54.78   &42.46   &60.12\\
    & & w/o Sup-SCL  &80.16&52.20&63.01 &93.78  &76.63  &82.38  &72.10   &61.56  &73.13   \\
    & & Full  & \textbf{81.94} & \textbf{56.53} & \textbf{65.85} & \textbf{94.17} & \textbf{77.95} & \textbf{83.12}&
    \textbf{74.91}&\textbf{65.45}&\textbf{75.76}\\
    \midrule
    \midrule
     \specialrule{0em}{1pt}{1pt}
    \multirow{6}{*}[-1.5ex]{{50\%}} &
    w/o Semi-SCL & Full & 83.83  &59.90  &68.63  &94.52   &79.14   &84.28   &76.74   &69.80   &80.36\\
    \cline{2-12}
    \specialrule{0em}{1pt}{1pt}
        &\multirow{5}{*}{Full} 
    & K-Means++&73.60&45.48&58.95&87.33&62.64&73.60&58.08&44.72&64.91\\
    & & w/o Self-Sup  & 77.20  & 51.28  & 64.32  &92.59   &75.72   &83.42  &68.01   &51.64   &72.40\\
     & & w/o CGM  &  79.67 & 55.32  & 67.18 &92.95   &76.47   &83.41  &67.19   &58.64   &75.02\\
    & & w/o Sup-SCL  &83.54&59.87&69.39  &94.67 &80.34  &85.64   &77.18   &68.79   &80.00\\
    & & Full  & \textbf{85.05} & \textbf{63.77} & \textbf{73.27} & \textbf{95.48} & \textbf{82.99} & \textbf{87.28}&
\textbf{78.77}&\textbf{71.63}&\textbf{82.06} \\
    \midrule
    \midrule
     \specialrule{0em}{1pt}{1pt}
    \multirow{6}{*}[-1.5ex]{{75\%}} &
    w/o Semi-SCL & Full &86.70  &67.32  &75.91  &96.10   &85.14   &88.99   &79.03   &73.78   &84.17\\
    \cline{2-12}
    \specialrule{0em}{1pt}{1pt}
        &\multirow{5}{*}{Full} 
    & K-Means++ &78.06&53.89&67.29&90.24&70.05&79.29&68.10&54.93&74.78 \\
    & & w/o Self-Sup  & 81.80  & 60.33  & 72.60  &94.84   &82.79   &88.41  &73.57   &57.51   &78.00\\
     & & w/o CGM  &  83.65 & 63.52	  & 74.51 &95.07  &83.14   &88.23  &74.78   &67.75  &81.87\\
    & & w/o Sup-SCL  &85.91&66.05&75.61  &95.70  &84.22  &88.86   &78.58   &72.06   &83.52 \\
    & & Full  & \textbf{87.41} & \textbf{69.54} & \textbf{78.36} & \textbf{96.42} & \textbf{86.77} & \textbf{90.36}&
    \textbf{80.13}&\textbf{74.90}&\textbf{85.66} \\
    \bottomrule
\end{tabular}}
\end{table*}
\begin{table}\tiny
\caption{
Cluster number estimation results in unsupervised and semi-supervised settings on the three datasets.
}
\resizebox{1.\columnwidth}{!}{
\begin{tabular}{@{\extracolsep{2pt}}clcccccc}
    \toprule
    \centering
    &  & \multicolumn{2}{c}{BANKING} 
    & \multicolumn{2}{c}{CLINC150} 
    & \multicolumn{2}{c}{StackOverflow}\\
    \addlinespace[0.05cm] \cline{3-4} \cline{5-6}  \cline{7-8}
    \addlinespace[0.05cm]
    KCR & Methods &  $K$ & Error& $K$ & Error & $K$ & Error\\
    \midrule
    {0\%}
    & USNID &\textbf{74.00}&\textbf{3.90}&\textbf{137.80}&\textbf{8.13}&\textbf{16.80}&\textbf{16.00}	\\
    \midrule
    \midrule
    \multirow{1.5}{*}[-1.5ex]{25\%}
    & DTC & 42.30  & 45.06    & 108.20   & 27.87   &9.50  &52.50     \\
    & DeepAligned &63.50  &17.53   &122.00  &18.67  &16.60 &17.00    \\
    & USNID &\textbf{74.30}&\textbf{3.51}&\textbf{139.60}&\textbf{6.93}&\textbf{16.80}&\textbf{16.00}	\\
    \midrule	
  \midrule
    \multirow{1.5}{*}[-1.5ex]{ 50\%}
    & DTC &83.40  &8.31    &157.50  &5.00     &\textbf{18.90}  &\textbf{5.50}   \\
    & DeepAligned &65.10  &15.45   &125.60  &16.27  &11.40  &43.00 \\
    & USNID  & \textbf{77.50} &\textbf{0.65}&\textbf{143.20} &\textbf{4.53}&18.30 &8.50	\\
    \midrule
\midrule
    \multirow{1.5}{*}[-1.5ex]{75\%}
    & DTC          &112.00  &45.45   &218.00  &45.33    &27.10   &35.50      \\
    & DeepAligned  &68.83  &10.61   &128.60  &14.27   &16.60  &17.00 \\
    & USNID &\textbf{82.80}&\textbf{7.53}&\textbf{154.50}&\textbf{3.00}&\textbf{18.90}&\textbf{5.50}	\\
    \bottomrule
\end{tabular}}
\label{cluster-number-est} 
\end{table}

\section{Results and Discussion}
\subsection{Results of New Intent Discovery}
The main experimental results of unsupervised and semi-supervised new intent discovery are presented in Table~\ref{results-main}. We highlight the best results for each setting (KCR=0\%, 25\%, 50\%, and 75\%) in bold and conduct significance $t$ tests between our method (USNID) and the other baselines\footnote{Since DCSC is not open source, we only report the results as in~\cite{wei2022semi}.}. 

In unsupervised new intent discovery (KCR=0\%), traditional clustering methods (e.g., KM and AG) show the lowest performance across all datasets, mainly due to their inability to comprehend complicated semantics using static feature-engineering representations. On the contrary, deep clustering methods demonstrate superior performance (over 10\% score improvement on ARI on BANKING and StackOverflow datasets) by learning representations end-to-end with deep neural networks during clustering. While CC incorporates instance-level and cluster-level contrastive learning techniques, its limitations in capturing cluster-level relations without learning specific targets lead to performance inferior to even some traditional methods. SCCL, the current SOTA unsupervised method in NLP, improves performance by replacing cluster-level contrastive learning with a technique forcing each sample to learn from high-confidence constructed soft targets via KL-divergence. Yet, our method, USNID, outperforms SCCL by 16.35\%, 30.14\%, and 17.44\% in ARI scores on the BANKING, CLINC150, and StackOverflow datasets, respectively, and shows improvements of over 10\% in all three metrics on the BANKING and CLINC150 datasets.


In semi-supervised new intent discovery (KCR$\neq$0\%), various methods (e.g., KCL, MCL, GCD, and CDAC+) construct pairwise similarity relations and use them to learn friendly representations for clustering by pulling similar and repelling dissimilar pairs. DTC constructs the target distribution as in SCCL and extends it by incorporating temporal information and consistent constraints. Despite this, their pairwise constraints have weak correlations, resulting in difficulties in distinguishing complex semantic intent-wise groups. In contrast, DeepAligned and DCSC, which use alignment strategies to generate categorical discrimination pseudo-labels, show significant improvements of over 10\% scores in ARI. MTP-CLNN is the existing  SOTA method in the semi-supervised setting, intensifies the pairwise constraints by including additional similarity connections in the nearest neighbor space. However, this could lead to unstable clustering targets as the representation learning process progresses. Our method, USNID, uses relatively stable targets guided by cluster centroids and achieves better results. Notably, USNID outperforms MTP-CLNN in ARI by 10.65\%, 9.39\%, and 5.54\% with 25\%, 50\%, and 75\% known classes on the StackOverflow dataset, respectively and shows significant improvements in NMI and ARI by over 1\% across nearly all settings.



\begin{figure*}[t!]\small
\centering
\includegraphics[width=2.05\columnwidth]{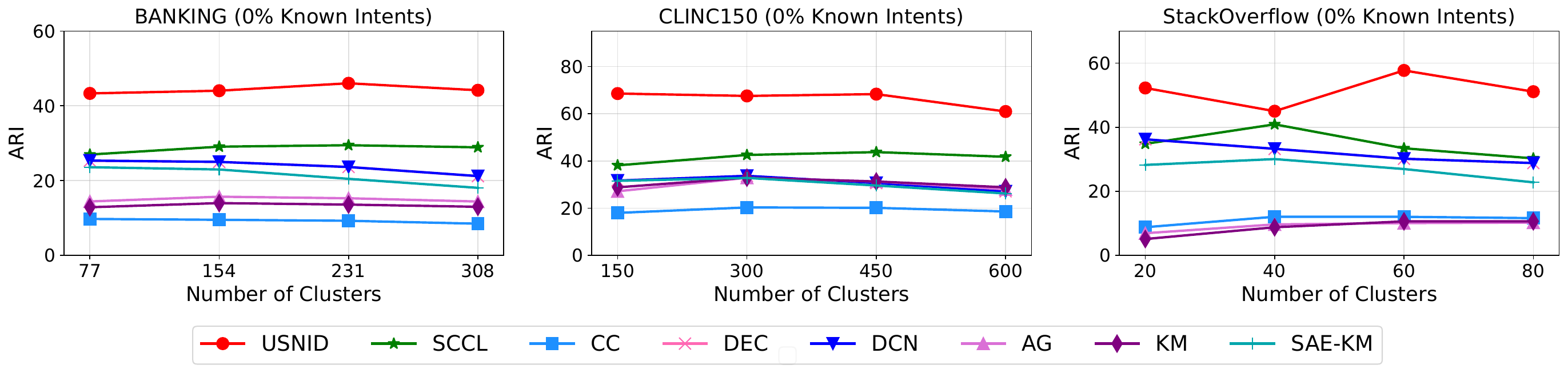}
\scalebox{1}{\includegraphics[width=2.05\columnwidth]{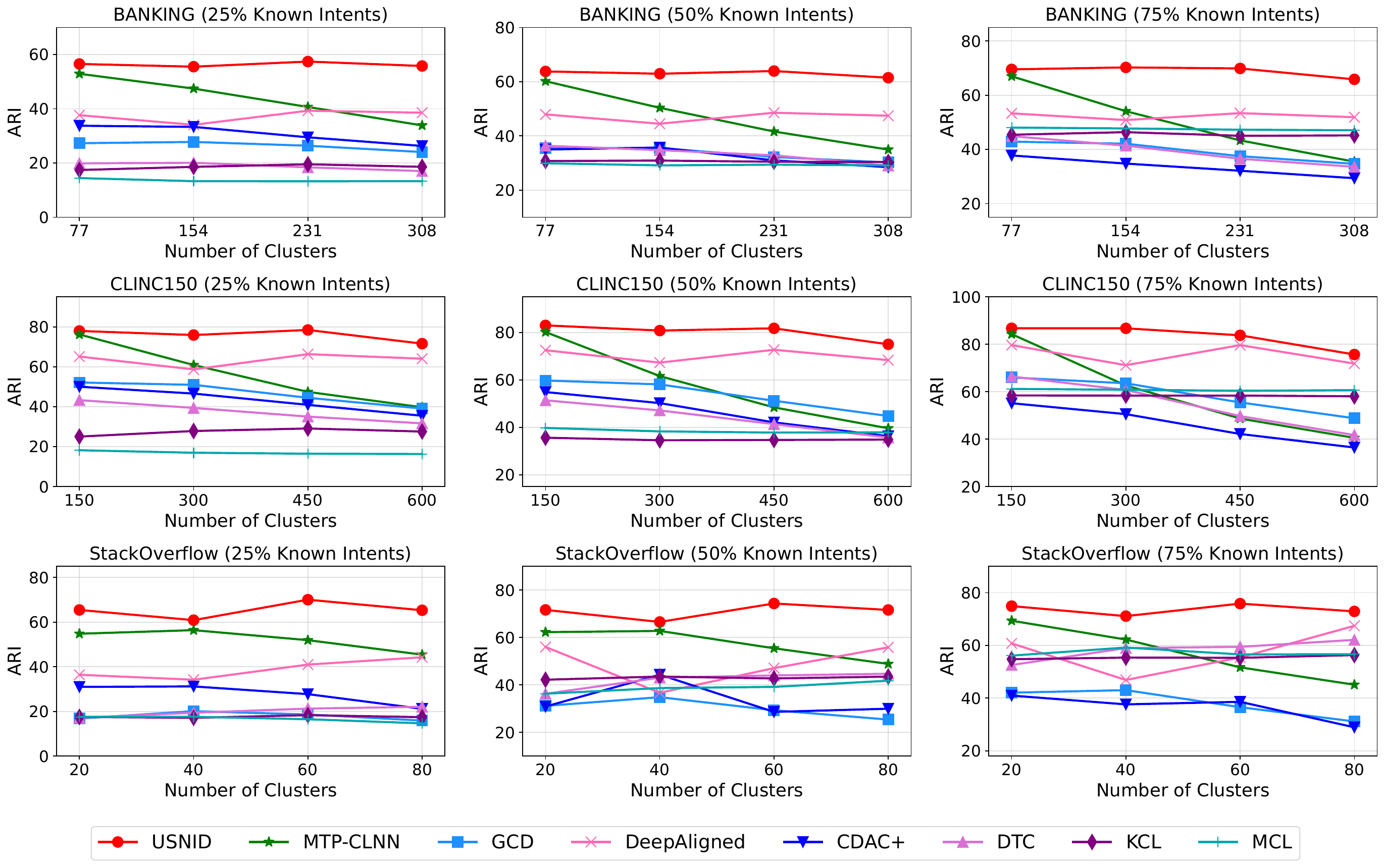}}
\caption{\label{cluster-number}  Unsupervised and semi-supervised new intent discovery results with different cluster numbers on the three datasets.}
\end{figure*}

Interestingly, the performance of unsupervised USNID outperforms more than half of the semi-supervised clustering methods with 75\% known classes. This is attributed to two main factors: (1) USNID, unlike other methods only using limited labeled data for pre-training, it strongly augments all samples and applies unsupervised contrastive learning, providing superiorly initialized representations and avoiding potential overfitting problem. (2) It uses a centroid-guided mechanism to create specific pseudo-labels rather than ambiguous pairwise relations as clustering targets, which brings categorical information to guide class differentiation.



The clustering performance is notably superior on the CLINC150 dataset compared to the BANKING dataset, primarily due to their inherent differences. The CLINC150 dataset contains utterances from 10 general domains, enabling better distinction of intent classes. In contrast, the BANKING dataset, originates from a singular banking domain, limiting the use of diverse semantic backgrounds for intent detection. Additionally, the BANKING dataset presents overlapping intent categories and contains longer text sequences with complex semantics, thus posing increased challenges to the model's capability to generalize effectively. The performance improves as the number of known classes increases, indicating the positive influence of labeled data on clustering. Semi-supervised USNID achieves top-tier results across all settings and substantial improvements over all baselines and unsupervised USNID, highlighting the advantage of our method in utilizing labeled data for new intent discovery.

\subsection{Ablation Studies}
\label{ablation}
To validate the effectiveness of the components in USNID, we conduct comprehensive ablation studies and show the results in Table~\ref{results-ablation}. USNID has two stages: pre-training (stage 1) and clustering and representation learning (stage 2). For stage 1, removing the unsupervised contrastive loss (\textit{w/o UCL}) results in a 13-34\% drop in ARI, while removing the semi-supervised contrastive loss (\textit{w/o Semi-SCL}) causes a 1-4\% decrease across three datasets. These results highlight the importance of pre-training in generating well-initialized representations for clustering. In stage 2, directly performing K-Means++ after stage 1 causes a 15-43\% and 15-31\% absolute ARI decrease in unsupervised and semi-supervised settings, respectively. This underscores the significance of this stage. Without the centroid-guided mechanism (\textit{w/o CGM}), performing K-Means++ once and using its pseudo-labels as targets for representation learning leads   This suggests that using historical centroids as guidance for updating pseudo-labels effectively constructs high-quality self-supervised signals for feature learning. Furthermore, removing the cross-entropy loss (\textit{w/o CE}) in the unsupervised setting leads to a 2-30\% decrease across all datasets. In the semi-supervised setting, removing the self-supervised learning loss (\textit{w/o Self-Sup}) results in decreases of 5-15\%, 1-10\%, and 6-22\% in various KCR settings across all datasets. This implies that specific pseudo-labels significantly improve clustering performance over the pairwise constraints of the contrastive loss. Lastly, omitting the additional supervised contrastive loss (\textit{w/o Sup-SCL}) in the semi-supervised setting leads to a 1-4\% drop in ARI across all KCR settings of the three datasets. This shows that \textit{Sup-SCL} mitigates the \textit{catastrophic forgetting} problem and better uses labeled data.

\begin{figure*}[t!]
\centering\includegraphics[width=2.05\columnwidth ]
{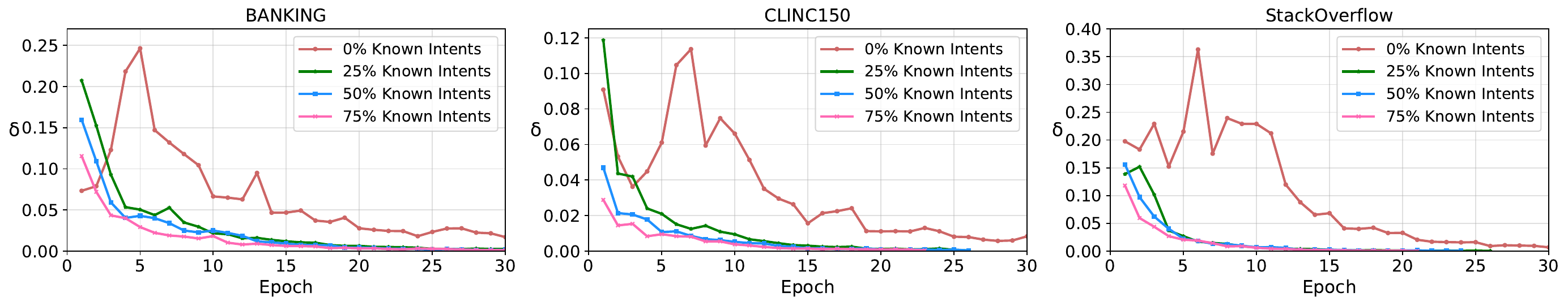}
    \caption{\label{curve}  The convergence curves of USNID in unsupervised and semi-supervised settings on the three datasets.}
\end{figure*}


\subsection{Cluster Number Estimation}
In this section, we explore new intent discovery in a more challenging situation where the ground truth number of clusters is not known in advance. As suggested in~\cite{Zhang_Xu_Lin_Lyu_2021}, the initial cluster number is set to large values (i.e., twice the ground truth number) of 154, 300, and 40 for BANKING, CLINC150, and StackOverflow, respectively. We compare our approach (described in section~\ref{est-k}) with two strong baselines for estimating the number of clusters, as proposed in~\cite{Han2019learning,Zhang_Xu_Lin_Lyu_2021}. To evaluate the accuracy of the estimated cluster numbers, we compute the error between the average of the estimated $K$ (obtained from ten runs of experiments) and the ground truth number, with the lower error being better. The results are shown in Table~\ref{cluster-number-est}. 

Our method consistently achieves the lowest errors on all three datasets in semi-supervised settings, with the exception of the 50\% KCR setting on the StackOverflow dataset. Though DTC performs well on the 50\% KCR setting, it is unstable with different amounts of labeled data and performs worse than the other two methods in the 25\% and 75\% settings on all three datasets. DeepAligned is a preliminary version of our method that also predicts $K$ by removing low-confidence clusters. However, it ignores the use of prior knowledge of labeled data to induce the known intent clusters, which may be falsely dropped under the assumption of high-quality cluster selection. As a result, it usually yields lower prediction results with 4-14\%, 11-17\%, and 1-34\% higher errors than our method on the three datasets for all KCR settings. Interestingly, USNID even exhibits competitive performance in the unsupervised setting, suggesting that it can also benefit from the weak semantic similarity relations learned through unsupervised contrastive learning. 

\subsection{Effect of the Number of Clusters}
In this section, we study the impact of the cluster number on the performance of new intent discovery. As suggested in~\cite{lin2020discovering,Zhang_Xu_Lin_Lyu_2021}, we vary the number of clusters in the range of one to four times the ground truth number. The last three settings correspond to open-world scenarios for discovering new intents, as is often the case in real applications. We use ARI as the  metric and show the results in Figure~\ref{cluster-number}. 

In unsupervised new intent discovery, USNID consistently demonstrates the best performance with significant improvements over the other methods on all three datasets. It also maintains robust performance with small fluctuations even when the assigned cluster number is large. We observe that most unsupervised baselines are not sensitive to the number of clusters, especially on the BANKING and CLINC150 datasets. However, their performance remains poor compared to the ground truth number, and there is still a significant gap between them and our method.

In semi-supervised new intent discovery, USNID also outperforms all other baselines on the three datasets. It is particularly more robust than other methods, and its performance is only slightly affected by the number of clusters. In contrast, while MTP-CLNN performs well when using the ground truth number, it is extremely sensitive to the cluster number. Its performance drops dramatically as the cluster number increases, resulting in much lower performance than USNID. DeepAligned is relatively more robust among these baselines as it uses a similar strategy to estimate the cluster number as our method. 

\subsection{Convergence Analysis}
In this section, we analyze the convergence of USNID by tracking the variation of the cluster allocation difference $\delta$ (described in section~\ref{c-g-c}) over the number of training epochs. The results are depicted in Figure~\ref{curve}.

It can be observed that, when using labeled data as prior knowledge, our method stably and efficiently converges to a small threshold (i.e., 0.0005) within a few epochs (around 25) on all three datasets. This demonstrates the usefulness of labeled data in guiding the clustering performance. Furthermore, increasing the amount of labeled data from 25\% to 75\% of known intents leads to a lower $\delta$ and generally faster convergence time. In the more challenging unsupervised setting, although there are fluctuations in the first few epochs, our method is still able to converge gradually to a small value. We suggest that this is because the provided aligned targets, although not guided by any prior knowledge, can still produce high-quality, consistent targets despite potentially introducing some noise.

\section{Conclusions and Future Work}
In this paper, we address the problem of new intent discovery in both unsupervised and semi-supervised settings, which is relevant to real-world scenarios. We propose USNID, a novel clustering framework that utilizes several key techniques to tackle this problem. First, USNID captures elementary semantic features through a pre-training stage by learning similarity information with self-augmented samples or limited labeled data, which has been shown to enhance the subsequent clustering. Second, we introduce a novel centroid-guided clustering mechanism to address the issue of inconsistent cluster assignments in partition-based methods during multiple clustering. This method obtains aligned targets by initializing the current clustering with the cluster centroids from the previous clustering, which demonstrates efficient convergence. Third, USNID learns fine-grained intent-specific group characteristics by jointly learning cluster-level and instance-level information with the targets of aligned pseudo-labels from the previous iteration's clustering, which significantly improves performance with clustering-friendly representations. Incorporating high-quality prior knowledge from labeled data has also been shown to bring additional benefits. When evaluated on several intent benchmarks, USNID outperforms all other methods in unsupervised and semi-supervised settings by a significant margin. Furthermore, we propose an effective method for estimating the number of clusters, which helps maintain robust performance in realistic scenarios without prior knowledge of the number of new classes.

In this study, our approach still depends on the specification of a large cluster count, an aspect that is somewhat reliant on empirical experience. In our future research, we intend to investigate the possibility of automatically determining the number of clusters with minimal assumptions. Additionally, our proposed framework is currently offline and necessitates the completion of clustering across all data. We foresee the potential for future research to extend our methodology to an online strategy, which could prove more efficient and applicable to larger datasets. Finally, while we currently employ K-Means++, there might be potential for exploring more efficient and effective centroid-based clustering methods. By addressing these areas for improvement, we aspire to expand the capabilities of our framework, ultimately enhancing both its applicability and efficiency.


	
	%

	\ifCLASSOPTIONcompsoc

	\ifCLASSOPTIONcaptionsoff
	\newpage
	\fi
	
	\bibliographystyle{IEEEtran}
	\bibliography{TKDE}

\begin{thebibliography}{10}
\providecommand{\url}[1]{#1}
\csname url@samestyle\endcsname
\providecommand{\newblock}{\relax}
\providecommand{\bibinfo}[2]{#2}
\providecommand{\BIBentrySTDinterwordspacing}{\spaceskip=0pt\relax}
\providecommand{\BIBentryALTinterwordstretchfactor}{4}
\providecommand{\BIBentryALTinterwordspacing}{\spaceskip=\fontdimen2\font plus
\BIBentryALTinterwordstretchfactor\fontdimen3\font minus \fontdimen4\font\relax}
\providecommand{\BIBforeignlanguage}[2]{{%
\expandafter\ifx\csname l@#1\endcsname\relax
\typeout{** WARNING: IEEEtran.bst: No hyphenation pattern has been}%
\typeout{** loaded for the language `#1'. Using the pattern for}%
\typeout{** the default language instead.}%
\else
\language=\csname l@#1\endcsname
\fi
#2}}
\providecommand{\BIBdecl}{\relax}
\BIBdecl

\bibitem{ijcai2021p622}
L.~Qin, T.~Xie, W.~Che, and T.~Liu, ``A survey on spoken language understanding: Recent advances and new frontiers,'' in \emph{Proc. 30th Int. Joint Conf. Artif. Intell.}, 2021, pp. 4577--4584.

\bibitem{li2022automatically}
Y.~Li, C.~Gao, X.~Du, H.~Wei, H.~Luo, D.~Jin, and Y.~Li, ``Automatically discovering user consumption intents in meituan,'' in \emph{Proc. 28th ACM SIGKDD Int. Conf. Knowl. Discovery Data Mining}, 2022, pp. 3259--3269.

\bibitem{lin2020discovering}
T.-E. Lin, H.~Xu, and H.~Zhang, ``Discovering new intents via constrained deep adaptive clustering with cluster refinement,'' in \emph{Proc. 34th AAAI Conf. Artif. Intell.}, 2020, pp. 8360--8367.

\bibitem{9319534}
H.~Li, X.~Wang, Z.~Zhang, J.~Ma, P.~Cui, and W.~Zhu, ``Intention-aware sequential recommendation with structured intent transition,'' \emph{IEEE Trans. Knowl. Data Eng.}, pp. 5403--5414, 2022.

\bibitem{schuurmans2019intent}
J.~Schuurmans and F.~Frasincar, ``Intent classification for dialogue utterances,'' \emph{IEEE Trans. Intell. Transp. Syst.}, pp. 82--88, 2019.

\bibitem{han2021autonovel}
K.~Han, S.-A. Rebuffi, S.~Ehrhardt, A.~Vedaldi, and A.~Zisserman, ``Autonovel: Automatically discovering and learning novel visual categories,'' \emph{IEEE Trans. Pattern Anal. Mach. Intell.}, pp. 6767--6781, 2021.

\bibitem{fini2021unified}
E.~Fini, E.~Sangineto, S.~Lathuili{\`e}re, Z.~Zhong, M.~Nabi, and E.~Ricci, ``A unified objective for novel class discovery,'' in \emph{Proc. IEEE Int. Conf. Comput. Vis.}, 2021, pp. 9284--9292.

\bibitem{vaze2022generalized}
S.~Vaze, K.~Han, A.~Vedaldi, and A.~Zisserman, ``Generalized category discovery,'' in \emph{Proc. IEEE Conf. Comput. Vis. Pattern Recognit.}, 2022, pp. 7492--7501.

\bibitem{Han2019learning}
K.~Han, A.~Vedaldi, and A.~Zisserman, ``Learning to discover novel visual categories via deep transfer clustering,'' in \emph{Proc. IEEE Int. Conf. Comput. Vis.}, 2019, pp. 8400--8408.

\bibitem{Zhang_Xu_Lin_Lyu_2021}
H.~Zhang, H.~Xu, T.-E. Lin, and R.~Lyu, ``Discovering new intents with deep aligned clustering,'' in \emph{Proc. AAAI Conf. Artif. Intell.}, 2021, pp. 14\,365--14\,373.

\bibitem{kumar-etal-2022-intent}
R.~Kumar, M.~Patidar, V.~Varshney, L.~Vig, and G.~Shroff, ``Intent detection and discovery from user logs via deep semi-supervised contrastive clustering,'' in \emph{Proc. Conf. North Amer. Chapter Assoc. Comput. Linguistics: Hum. Lang. Technol.}, 2022, pp. 1836--1853.

\bibitem{wei2022semi}
F.~Wei, Z.~Chen, Z.~Hao, F.~Yang, H.~Wei, B.~Han, and S.~Guo, ``Semi-supervised clustering with contrastive learning for discovering new intents,'' \emph{arXiv: 2201.07604}, 2022.

\bibitem{zhang2022new}
Y.~Zhang, H.~Zhang, L.-M. Zhan, X.-M. Wu, and A.~Lam, ``New intent discovery with pre-training and contrastive learning,'' in \emph{Proc. 60th Assoc. Comput. Linguistics}, 2022, pp. 256--269.

\bibitem{macqueen1967some}
J.~MacQueen \emph{et~al.}, ``Some methods for classification and analysis of multivariate observations,'' in \emph{Proc. 5th Berkeley Symp. Math. Statist. Probability.}, 1967, pp. 281--297.

\bibitem{gowda1978agglomerative}
K.~C. Gowda and G.~Krishna, ``Agglomerative clustering using the concept of mutual nearest neighbourhood,'' \emph{Pattern Recognit.}, pp. 105--112, 1978.

\bibitem{jain1999data}
A.~K. Jain, M.~N. Murty, and P.~J. Flynn, ``Data clustering: a review,'' \emph{ACM Comput. Surv.}, pp. 264--323, 1999.

\bibitem{ruspini1969new}
E.~H. Ruspini, ``A new approach to clustering,'' \emph{Inf. Technol. Control.}, pp. 22--32, 1969.

\bibitem{arthur2007k}
D.~Arthur and S.~Vassilvitskii, ``k-means++: the advantages of careful seeding,'' in \emph{Proc. SIAM Int. Conf. Data Mining}, 2007, pp. 1027--1035.

\bibitem{lecun2015deep}
Y.~LeCun, Y.~Bengio, and G.~Hinton, ``Deep learning,'' \emph{Nature}, pp. 436--444, 2015.

\bibitem{ren2022deep}
Y.~Ren, J.~Pu, Z.~Yang, J.~Xu, G.~Li, X.~Pu, P.~S. Yu, and L.~He, ``Deep clustering: A comprehensive survey,'' \emph{arXiv: 2210.04142}, 2022.

\bibitem{xie2016unsupervised}
J.~Xie, R.~Girshick, and A.~Farhadi, ``Unsupervised deep embedding for clustering analysis,'' in \emph{Proc. Int. Conf. Mach. Learn.}, 2016, pp. 478--487.

\bibitem{vincent2010stacked}
P.~Vincent, H.~Larochelle, I.~Lajoie, Y.~Bengio, P.-A. Manzagol, and L.~Bottou, ``Stacked denoising autoencoders: Learning useful representations in a deep network with a local denoising criterion.'' \emph{J. Mach. Learn. Res.}, pp. 3371--3408, 2010.

\bibitem{yang2017towards}
B.~Yang, X.~Fu, N.~D. Sidiropoulos, and M.~Hong, ``Towards k-means-friendly spaces: Simultaneous deep learning and clustering,'' in \emph{Proc. Int. Conf. Mach. Learn.}, 2017, pp. 3861--3870.

\bibitem{chang2017deep}
J.~Chang, L.~Wang, G.~Meng, S.~Xiang, and C.~Pan, ``Deep adaptive image clustering,'' in \emph{Proc. IEEE Int. Conf. Comput. Vis.}, 2017, pp. 5879--5887.

\bibitem{caron2018deep}
M.~Caron, P.~Bojanowski, A.~Joulin, and M.~Douze, ``Deep clustering for unsupervised learning of visual features,'' in \emph{Proc. Eur. Conf. Comput. Vis.}, 2018, pp. 132--149.

\bibitem{chen2020simple}
T.~Chen, S.~Kornblith, M.~Norouzi, and G.~Hinton, ``A simple framework for contrastive learning of visual representations,'' in \emph{Proc. Int. Conf. Mach. Learn.}, 2020, pp. 1597--1607.

\bibitem{10.5555/1404506}
S.~Basu, I.~Davidson, and K.~Wagstaff, \emph{Constrained Clustering: Advances in Algorithms, Theory, and Applications}.\hskip 1em plus 0.5em minus 0.4em\relax USA:CRC Press, 2008.

\bibitem{wagstaff2001constrained}
K.~Wagstaff, C.~Cardie, S.~Rogers, and S.~Schr{\"o}dl, ``Constrained k-means clustering with background knowledge,'' in \emph{Proc. Int. Conf. Mach. Learn.}, 2001, pp. 577--584.

\bibitem{basu2004active}
S.~Basu, A.~Banerjee, and R.~J. Mooney, ``Active semi-supervision for pairwise constrained clustering,'' in \emph{Proc. SIAM Int. Conf. Data Mining}.\hskip 1em plus 0.5em minus 0.4em\relax SIAM, 2004, pp. 333--344.

\bibitem{bilenko2004integrating}
M.~Bilenko, S.~Basu, and R.~J. Mooney, ``Integrating constraints and metric learning in semi-supervised clustering,'' in \emph{Proc. Int. Conf. Mach. Learn.}, 2004, pp. 81--88.

\bibitem{hsu2018learning}
Y.-C. Hsu, Z.~Lv, and Z.~Kira, ``Learning to cluster in order to transfer across domains and tasks,'' in \emph{Proc. Int. Conf. Learn. Representations}, 2018.

\bibitem{hsu2018multiclass}
Y.-C. Hsu, Z.~Lv, J.~Schlosser, P.~Odom, and Z.~Kira, ``Multi-class classification without multi-class labels,'' in \emph{Proc. Int. Conf. Learn. Representations}, 2019.

\bibitem{YuSLCHH19}
E.~Yu, J.~Sun, J.~Li, X.~Chang, X.-H. Han, and A.~G. Hauptmann, ``Adaptive semi-supervised feature selection for cross-modal retrieval,'' \emph{IEEE Trans. Multim.}, vol.~21, no.~5, pp. 1276--1288, 2019.

\bibitem{YuanCLH22}
D.~Yuan, X.~Chang, Z.~Li, and Z.~He, ``Learning adaptive spatial-temporal context-aware correlation filters for uav tracking,'' \emph{ACM Trans. Multimedia Comput. Commun. Appl.}, vol.~18, no.~3, 2022.

\bibitem{han2019automatically}
K.~Han, S.-A. Rebuffi, S.~Ehrhardt, A.~Vedaldi, and A.~Zisserman, ``Automatically discovering and learning new visual categories with ranking statistics,'' in \emph{Proc. Int. Conf. Learn. Representations}, 2020.

\bibitem{Casanueva2020}
I.~Casanueva, T.~Temcinas, D.~Gerz, M.~Henderson, and I.~Vulic, ``Efficient intent detection with dual sentence encoders,'' \emph{arXiv: 2003.04807}, 2020.

\bibitem{zhang2022mintrec}
H.~Zhang, H.~Xu, X.~Wang, Q.~Zhou, S.~Zhao, and J.~Teng, ``Mintrec: A new dataset for multimodal intent recognition,'' in \emph{Proc. of the 30th ACM Int. Conf. on Multimedia}, 2022, pp. 1688--1697.

\bibitem{Zhang_Xu_Lin_2021}
H.~Zhang, H.~Xu, and T.-E. Lin, ``Deep open intent classification with adaptive decision boundary,'' in \emph{Proc. 35th AAAI Conf. Artif. Intell.}, 2021, pp. 14\,374--14\,382.

\bibitem{10097558}
H.~Zhang, H.~Xu, S.~Zhao, and Q.~Zhou, ``Learning discriminative representations and decision boundaries for open intent detection,'' \emph{IEEE/ACM Trans. Audio, Speech, and Lang. Process.}, vol.~31, pp. 1611--1623, 2023.

\bibitem{caron2020unsupervised}
M.~Caron, I.~Misra, J.~Mairal, P.~Goyal, P.~Bojanowski, and A.~Joulin, ``Unsupervised learning of visual features by contrasting cluster assignments,'' in \emph{Proc. Advances. Neural Inf. Proces. Syst.}, 2020, pp. 9912--9924.

\bibitem{cuturi2013sinkhorn}
M.~Cuturi, ``Sinkhorn distances: Lightspeed computation of optimal transport,'' \emph{Proc. Advances Neural Inf. Process. Syst.}, pp. 2292--2300, 2013.

\bibitem{larson-etal-2019-evaluation}
S.~Larson, A.~Mahendran, J.~J. Peper, C.~Clarke, A.~Lee, P.~Hill, J.~K. Kummerfeld, K.~Leach, M.~A. Laurenzano, L.~Tang, and J.~Mars, ``An evaluation dataset for intent classification and out-of-scope prediction,'' in \emph{Proc. Conf. Empir. Methods Natural Lang. Process.}, 2019, pp. 1311--1316.

\bibitem{van2020scan}
W.~Van~Gansbeke, S.~Vandenhende, S.~Georgoulis, M.~Proesmans, and L.~Van~Gool, ``Scan: Learning to classify images without labels,'' in \emph{Proc. Eur. Conf. Comput. Vis.}, 2020, pp. 268--285.

\bibitem{BERT}
J.~D. M.-W.~C. Kenton and L.~K. Toutanova, ``Bert: Pre-training of deep bidirectional transformers for language understanding,'' in \emph{Proc. North Amer. Chapter Assoc. Comput. Linguistics: Hum. Lang. Technol.}, 2019, pp. 4171--4186.

\bibitem{zhang2021supporting}
D.~Zhang, F.~Nan, X.~Wei, S.-W. Li, H.~Zhu, K.~R. McKeown, R.~Nallapati, A.~O. Arnold, and B.~Xiang, ``Supporting clustering with contrastive learning,'' in \emph{Proc. North Amer. Chapter Assoc. Comput. Linguistics: Hum. Lang. Technol.}, 2021, pp. 5419--5430.

\bibitem{zhan2020online}
X.~Zhan, J.~Xie, Z.~Liu, Y.-S. Ong, and C.~C. Loy, ``Online deep clustering for unsupervised representation learning,'' in \emph{Proc. IEEE Conf. Comput. Vis. Pattern Recognit.}, 2020, pp. 6688--6697.

\bibitem{kuhn1955hungarian}
H.~W. Kuhn, ``The hungarian method for the assignment problem,'' \emph{Nav. Res. Logistics Quart.}, pp. 83--97, 2010.

\bibitem{li2022twin}
Y.~Li, M.~Yang, D.~Peng, T.~Li, J.~Huang, and X.~Peng, ``Twin contrastive learning for online clustering,'' \emph{Int. J. Comput. Vis.}, vol. 130, no.~9, pp. 2205--2221, 2022.

\bibitem{khosla2020supervised}
P.~Khosla, P.~Teterwak, C.~Wang, A.~Sarna, Y.~Tian, P.~Isola, A.~Maschinot, C.~Liu, and D.~Krishnan, ``Supervised contrastive learning,'' in \emph{Proc. Advances Neural Inf. Process. Syst.}, 2020, pp. 18\,661--18\,673.

\bibitem{gao2021simcse}
T.~Gao, X.~Yao, and D.~Chen, ``Simcse: Simple contrastive learning of sentence embeddings,'' in \emph{Proc. Conf. Empir. Methods Natural Lang. Process.}, 2021, pp. 6894--6910.

\bibitem{xu-etal-2015-short}
J.~Xu, P.~Wang, G.~Tian, B.~Xu, J.~Zhao, F.~Wang, and H.~Hao, ``Short text clustering via convolutional neural networks,'' in \emph{Proc. North Amer. Chapter Assoc. Comput. Linguistics: Hum. Lang. Technol.}, 2015.

\bibitem{li2021contrastive}
Y.~Li, P.~Hu, Z.~Liu, D.~Peng, J.~T. Zhou, and X.~Peng, ``Contrastive clustering,'' in \emph{Proc. AAAI Conf. Artif. Intell.}, 2021, pp. 8547--8555.

\bibitem{pennington-etal-2014-glove}
J.~Pennington, R.~Socher, and C.~Manning, ``{G}lo{V}e: Global vectors for word representation,'' in \emph{Proc. Conf. Empir. Methods Natural Lang. Process.}, 2014, pp. 1532--1543.

\bibitem{reimers-gurevych-2019-sentence}
N.~Reimers and I.~Gurevych, ``Sentence-{BERT}: Sentence embeddings using {S}iamese {BERT}-networks,'' in \emph{Proc. Conf. Empir. Methods Natural Lang. Process.}, 2019, pp. 3982--3992.

\bibitem{wolf2020transformers}
T.~Wolf, L.~Debut, V.~Sanh, J.~Chaumond, C.~Delangue, A.~Moi, P.~Cistac, T.~Rault, R.~Louf, M.~Funtowicz \emph{et~al.}, ``Transformers: State-of-the-art natural language processing,'' in \emph{Proc. Conf. Empir. Methods Natural Lang. Process.}, 2020, pp. 38--45.

\bibitem{loshchilov2018decoupled}
I.~Loshchilov and F.~Hutter, ``Decoupled weight decay regularization,'' in \emph{Proc. Int. Conf. Learn. Representations}, 2019.

\bibitem{pedregosa2011scikit}
F.~Pedregosa, G.~Varoquaux, A.~Gramfort, V.~Michel, B.~Thirion, O.~Grisel, M.~Blondel, P.~Prettenhofer, R.~Weiss, V.~Dubourg \emph{et~al.}, ``Scikit-learn: Machine learning in python,'' \emph{J. Mach. Learn. Res.}, vol.~12, pp. 2825--2830, 2011.

\bibitem{zhang-etal-2021-textoir}
H.~Zhang, X.~Li, H.~Xu, P.~Zhang, K.~Zhao, and K.~Gao, ``{TEXTOIR}: An integrated and visualized platform for text open intent recognition,'' in \emph{Proc. 59th Assoc. Comput. Linguistics.}, 2021, pp. 167--174.

\end{thebibliography}

	
	
	%
	
	%

	\begin{IEEEbiography} 	[{\includegraphics[width=1in,height=1.1in,clip,keepaspectratio]{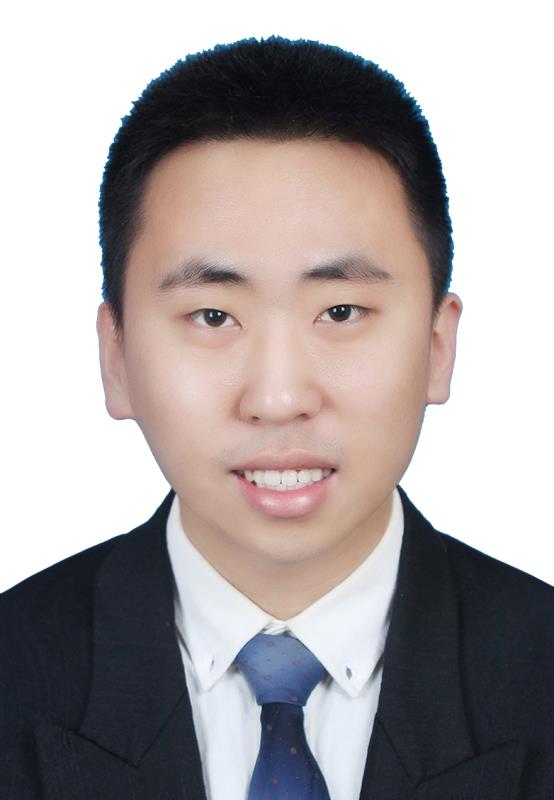}}]{Hanlei Zhang} received the B.S. degree from the
Department of Computer Science and Technology,
Beijing Jiaotong University, Beijing, China, in 2020. He is
currently working toward the Ph.D. degree with the
Department of Computer Science and Technology,
Tsinghua University, Beijing, China. He has authored or coauthored six papers in top-tier international conferences and journals, including AAAI, ACM MM, ACL, and IEEE/ACM Transactions on Audio, Speech and Language Processing. His research interests include intent analysis, open world classification, clustering, multimodal language understanding, and natural
language processing.
\end{IEEEbiography}
\begin{IEEEbiography}    [{\includegraphics[width=1in,height=1.25in,clip,keepaspectratio]{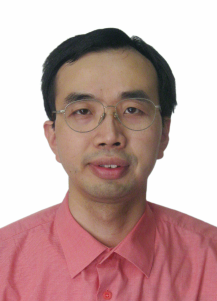}}]{Hua Xu}
     received the B.S. degree
from Xi’an Jiaotong University, Xi’an, China, in
1998, and the M.S. and Ph.D. degrees from Tsinghua
University, Beijing, China, in 2000 and 2003, respectively. He is a Tenured Associate Professor with the
Department of Computer Science and Technology,
Tsinghua University. He has authored or coauthored
more than 130 peer-reviewed papers in top-tier international journals and conferences. His research
interests include multi-modal intelligent information
processing for natural interaction of service robots,
evolutionary learning, and intelligent optimization. Prof. Xu was the recipient
of the Second Prize from the National Science and Technology Progress of
China, First Prize from Beijing Science and Technology, and Third Prize from
Chongqing Science and Technology.
\end{IEEEbiography}
\begin{IEEEbiography} 	[{\includegraphics[width=1in,height=1.1in,clip,keepaspectratio]{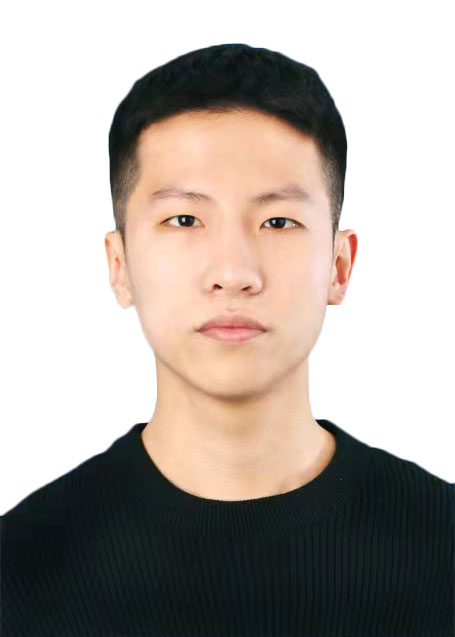}}]{Xin Wang} received the B.S. degree in 2020 from the School of Information Science and
Engineering, Hebei University of Science and Technology, Shijiazhuang, China, where he is currently working toward the M.S. degree with the School of Information Science and Engineering. He has authored or coauthored one paper in the ACM MM international conference. His research interests include  unsupervised learning, semi-supervised learning, and natural language processing. 
\end{IEEEbiography}
\begin{IEEEbiography} 	[{\includegraphics[width=1in,height=1in,clip,keepaspectratio]{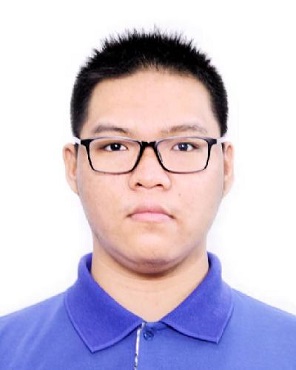}}]{Fei Long} is working toward the undergraduate degree in the Department of Computer
Science and Technology, Tsinghua University, Beijing, China. His current research interests include natural language processing, clustering, and machine learning.
\end{IEEEbiography}
\begin{IEEEbiography} 	[{\includegraphics[width=1in,height=1.1in,clip,keepaspectratio]{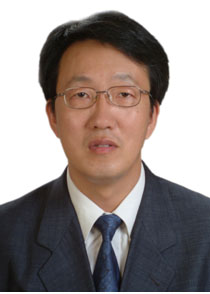}}]{Kai Gao}  received the Ph.D. degree from Shanghai Jiao Tong University, Shanghai, China. He is a Professor in the School of Information Science and Engineering, Hebei University of Science and Technology. He has authored or coauthored over 60 academic papers in international conferences and journals. His research interests
include natural language processing, knowledge discovery, and multimodal intelligent information processing. 
\end{IEEEbiography}
	
	

\end{document}